\documentclass{article}

\PassOptionsToPackage{numbers, compress}{natbib}

\usepackage[final]{neurips_2024}

\usepackage[utf8]{inputenc} 
\usepackage[T1]{fontenc}    
\usepackage{hyperref}       
\usepackage{url}            
\usepackage{booktabs}       
\usepackage{amsfonts}       
\usepackage{nicefrac}       
\usepackage{microtype}      
\usepackage[dvipsnames]{xcolor}         

\usepackage{subfigure}
\usepackage{graphicx}
\usepackage{amsmath}
\usepackage{algorithm}
\usepackage{algorithmic}
\usepackage{multirow, multicol}
\usepackage{adjustbox}
\usepackage{pifont}
\usepackage{bm}
\newcommand{\pub}[1]{{\color{gray}\tiny{[{#1}]}}}

\newcommand{\desalt}[1]{{\color{gray}{#1}}}

\title{Bridge the Points: Graph-based Few-shot \\ Segment Anything Semantically}

\author{%
  Anqi Zhang\textsuperscript{1}, Guangyu Gao\textsuperscript{1}\thanks{Corresponding Author.}, Jianbo Jiao\textsuperscript{2}, Chi Harold Liu\textsuperscript{1}, and Yunchao Wei\textsuperscript{3} \\
    \textsuperscript{1}School of Computer Science, Beijing Institute of Technology \\
    \textsuperscript{2}The MIx group, School of Computer Science, University of Birmingham \\
    \textsuperscript{3}WEI Lab, Institute of Information Science, Beijing Jiaotong University \\
    \texttt{andy\_zaq@outlook.com} \\
}

\begin{document}

\maketitle

\begin{abstract}
The recent advancements in large-scale pre-training techniques have significantly enhanced the capabilities of vision foundation models, notably the Segment Anything Model~(SAM), which can generate precise masks based on point and box prompts.
Recent studies extend SAM to Few-shot Semantic Segmentation~(FSS), focusing on prompt generation for SAM-based automatic semantic segmentation. 
However, these methods struggle with selecting suitable prompts, require specific hyperparameter settings for different scenarios, and experience prolonged one-shot inference times due to the overuse of SAM, resulting in low efficiency and limited automation ability.
To address these issues, we propose a simple yet effective approach based on graph analysis and representation learning. 
In particular, a Positive-Negative Alignment module dynamically selects the point prompts for generating masks, especially uncovering the potential of the background context as the negative reference.
Another subsequent Point-Mask Clustering module aligns the granularity of masks and selected points as a directed graph, based on mask coverage over points. 
These points are then aggregated by decomposing the weakly connected components of the directed graph in an efficient manner, constructing distinct natural clusters.
Finally, the positive and overshooting gating, benefiting from graph-based granularity alignment, aggregate high-confident masks and filter out the false-positive masks for final prediction, reducing the usage of additional hyperparameters and redundant mask generation. 
Extensive experimental analysis across standard FSS, One-shot Part Segmentation, and Cross Domain FSS datasets validate the effectiveness and efficiency of the proposed approach, surpassing state-of-the-art generalist models with a mIoU of 58.7\% on COCO-20\textsuperscript{i} and 35.2\% on LVIS-92\textsuperscript{i}. 
The project page of this work is \href{https://andyzaq.github.io/GF-SAM/}{https://andyzaq.github.io/GF-SAM/}.

\end{abstract}

\section{Introduction}\label{sec:intro}

Previous semantic segmentation methods~\cite{fcn,deeplabv1,deeplabv3,deeplabv2,segformer21,micro22,coinseg23,barm24}, which rely on the pixel-level classification, often struggle with generalization and overfitting due to limited labeled data. 
In addition, recent approaches, such as MaskFormer~\cite{maskformer}, have shifted the paradigm to mask-based classification, offering a more flexible approach to improving the segmentation performance by exploiting the consistency and completeness of generated class-agnostic masks. 
The Segment Anything Model~(SAM)~\cite{sam23} further marks a significant advancement by utilizing extensive pre-training on vast datasets SA-1B to achieve more robust, class-agnostic segmentation capabilities. 
SAM excels in producing precise masks across various domains using simple prompts such as points, boxes, and coarse masks. 
While the boundaries of these masks can closely align with object boundaries, the lack of semantic information and the requirement for manual prompts prevent SAM from being used in automatic semantic segmentation applications. 

\begin{figure}[t]
    \centering
    \subfigure[Performance-efficiency comparison of FSS models. The numbers inside the points represent the numbers of parameters.]{
        \label{fig:eff}        \includegraphics[width=0.54\linewidth]{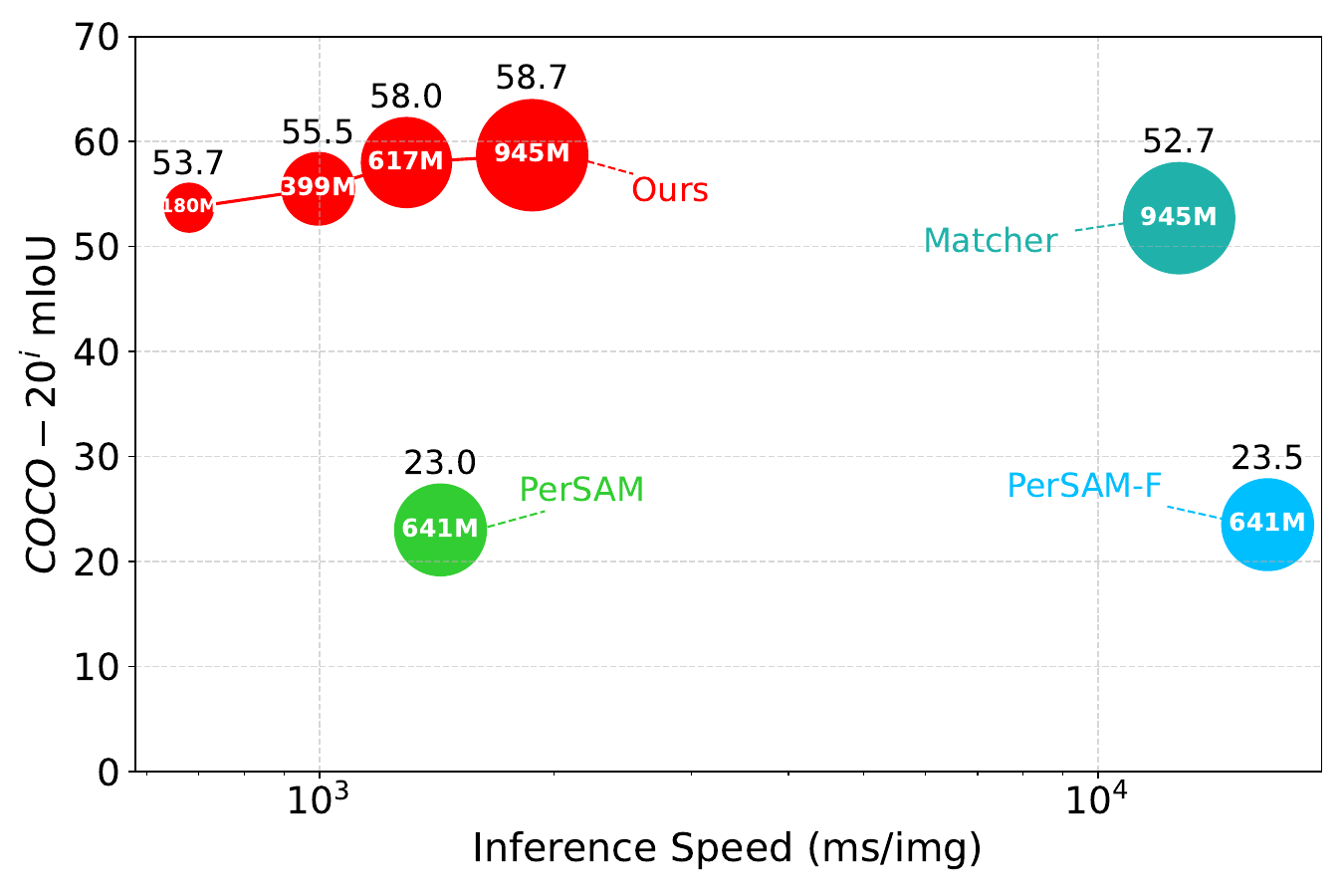}
    }
    \hfill
    \subfigure[Comparison with previous generalist and specialist models on various FSS datasets.]{
        \label{fig:radar}        \includegraphics[width=0.41\linewidth]{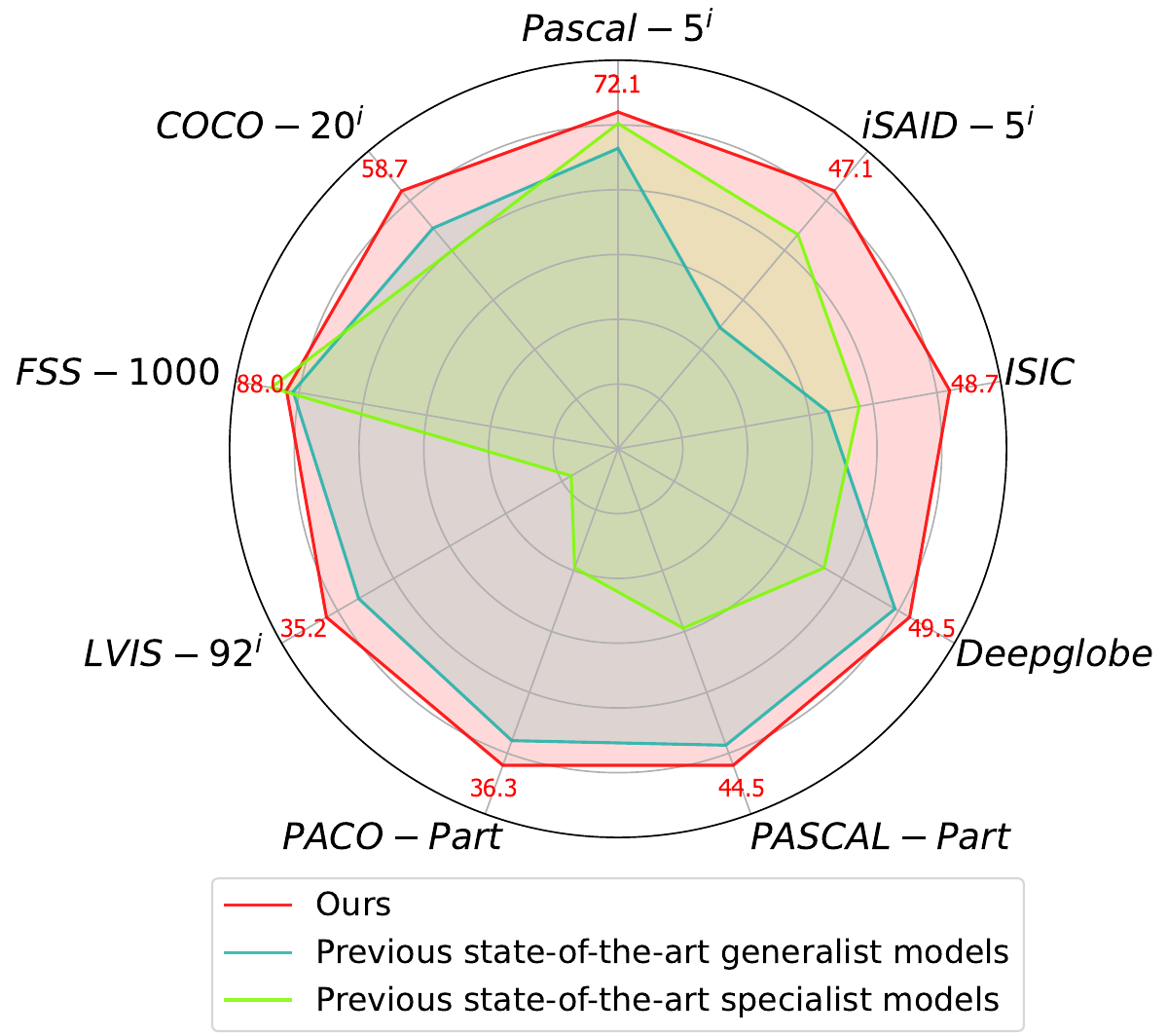}
    }
    \caption{Performance comparisons of our approach against previous state-of-the-art methods regarding efficiency and generalized capabilities in Few-shot Semantic Segmentation. 
    Figure 1(a) illustrates our approach's superior performance in efficiency and effectiveness across various model sizes. 
    Figure 1(b) demonstrates the generalizability of our approach across different domains.}
    \label{fig:perf}
\end{figure}

Recent studies have attempted to automate this process in the Few-shot Semantic Segmentation~(FSS), using a few reference images and a fine-grained external backbone network~(\textit{e.g.}, DINOv2~\cite{dinov223}) to guide SAM in segmenting target semantic objects. 
However, these methods face two main challenges: achieving suitable points for precise and full coverage of the target object, and handling the ambiguity of SAM-generated masks, from partial to complete coverage. 
Specifically, they either utilize the most similar candidate point prompts for iterative mask generation and refinement~\cite{persam24}, or build a restrictively selected set of point prompts for heuristically weighted mask merging based on manually designed metrics~\cite{matcher24}, outperforming both previous specialist methods~\cite{hsnet21,sccan,svf22,ntre22,hcnet23,drnet22} and generalist methods without SAM~\cite{seggpt23, painter23}. 
However, these methods overlooked the underlying relationships between points (derived from fine-grained features) and masks (generated by SAM in a coarse-grained manner). 
This oversight led to low efficiency (as indicated in Fig.~\ref{fig:eff}) and limited automation capabilities. 
Alignment between these two types of granularity could uncover the potential of simple decision-making on masks, which can eliminate redundant refinement and manual hyperparameter selection for complicated metrics. 

In this paper, we explicitly explore the relationship between point prompts and corresponding masks from SAM, and present a streamlined yet effective parameter-free framework with only one-time mask generation to segment anything in a graph-based few-shot manner semantically.
We first introduce a Positive-Negative Alignment~(PNA) module to dynamically select point prompts using foreground and background references from reference images. 
Unlike existing methods, our approach combines different granularity by constructing a directed graph according to mask coverage over points. 
Then, we perform connectivity analysis on the constructed graph to obtain several weakly connected components as automatic clustering of point prompts, 
which bridges points and masks as well as fine-grained and coarse-grained features. 
To mitigate the inevitable introduction of false positives in the PNA module, we further leverage weakly connected component clusters and limited semantic information in selected points, to more accurately filter and merge masks that mismatch in different granularities.
In particular, our proposed method involves two post-gating based on weakly connected clusters: the positive gating retains masks capturing a greater proportion of potential target areas, while the overshooting gating screens out outlier points near object boundary, with coverage self-consistency consideration.

Extensive experimental analysis on Few-shot Semantic Segmentation demonstrates both the efficiency and effectiveness of our approach, as shown in Fig.~\ref{fig:radar}. 
We first conduct the experiments on generalized FSS datasets, including Pascal-5\textsuperscript{i}~\cite{ShabanOne}, COCO-20\textsuperscript{i}~\cite{fwb19}, FSS-1000~\cite{fss1000} and LVIS-92\textsuperscript{i}~\cite{matcher24}. 
Our approach surpasses existing state-of-the-art approaches on these datasets, with $5.8\%$ and $2.2\%$ of improvement respectively on more challenging COCO-20\textsuperscript{i} and LVIS-92\textsuperscript{i}. 
As for the challenging One-shot Part Segmentation, our approach still exceeds previous methods with $1.6\%$ of mIoU on both PACO-Part and PASCAL-Part. 
Furthermore, to demonstrate the ability of our approach across different domains, we perform an evaluation on several specific datasets, including Deepglobe~\cite{deepglobe}, ISIC~\cite{isic18}, and iSAID-5\textsuperscript{i}~\cite{isaid521}. 
The proposed approach establishes new state-of-the-art performance on mIoU with $49.5\%$ on Deepglobe, $48.7\%$ on ISIC, and $47.3\%$ on iSAID-5\textsuperscript{i}. 

Overall, our contributions are summarized as follows:
\begin{itemize}
    \item We present, to our knowledge, the first graph-based approach for SAM-based few-shot semantic segmentation, modeling the relationship of SAM-generated masks in an automatic clustering manner.
    \item We propose a positive-negative alignment module and a post-gating strategy based on the weakly connected graph components, enabling a hyperparameter-free pipeline.
    \item Extensive experimental comparisons and analysis across several datasets over various settings show the effectiveness and efficiency of the proposed method.
\end{itemize}

\section{Related Work}


\textbf{Few-shot Semantic Segmentation.} 
Few-shot Semantic Segmentation~(FSS)~\cite{ShabanOne} aims to segment the target object using only a limited number of annotated reference samples for guidance. 
Previous FSS methods are mainly categorized into two types, namely the methods based on prototype matching~\cite{canet19,panet19,asgnet21,pfenet20,RePRI21,iprnet22,fptrans22} and methods based on pixel-wise matching~\cite{dcama22,fecanet23,abcnet,ipmt22,mmformer,cyctr21}. 
The methods based on prototype matching, \textit{e.g.} PFENet~\cite{pfenet20}, BAM~\cite{bam22}, SSP~\cite{ssp22}, use the Mask Average Pooling operation from SGOne~\cite{sgone20} to generate a prototype as a global representation of the reference features, and compare the target features with the prototypes. 
The methods based on pixel-wise matching compute the correlation of all pixels between target and reference features. 
Then different methods address the correlations through distinct mechanisms, such as 4D Convolution~(\textit{e.g.}, HSNet~\cite{hsnet21}) and Transformer~(\textit{e.g.}, HDMNet~\cite{hdmnet23}, AMFormer~\cite{amformer23}). 
Although these specialist models perform significantly on specific tasks, they are prone to overfitting the training samples and often struggle to adapt to domain shifts. 


\textbf{SAM-based Semantic Segmentation.} Recently, Segment Anything Model~(SAM)~\cite{sam23} has shown remarkable zero-shot class-agnostic segmentation capabilities using prompts like points, boxes, and coarse masks. 
However, the coarse-grained feature representation of SAM limits its effectiveness for fine-grained semantic segmentation tasks.
Several approaches have been proposed to extend SAM for semantic segmentation.
For example, Semantic-SAM~\cite{li2023semantic} jointly train the model on SA-1B and other semantic aware segmentation datasets to enhance granularity. 
OV-SAM~\cite{yuan2024ovsam} combines SAM and CLIP~\cite{clip21} for open-vocabulary semantic segmentation. 
Moreover, some methods introduce SAM into FSS tasks. 
PerSAM and PerSAM-F~\cite{persam24} leverage SAM for personalized segmentation with one-shot guidance. 
Matcher~\cite{matcher24} uses a SAM-based training-free structure, achieving impressive performance in both FSS and One-shot Part Segmentation. 
VRP-SAM~\cite{sun2024vrp} trains an external Visual Reference Prompt Encoder to automatically generate prompts from reference images using points, scribble, box, or masks. 
However, previous training-free methods struggled to balance performance and efficiency, often relying on excessive external manual hyperparameters.


\section{Preliminaries}


Few-shot Semantic Segmentation~(FSS) aims to segment target objects in an image with a few annotated reference images. 
Consider a scenario where each group of samples contains a target image $x^t$ and a reference image set $R = \{x^r_k, y^r_k\}_{k=1}^K$ with the size of $H\times W$, where $x^r_k$ and $y^r_k$ mean the $k_{th}$ reference image and its corresponding mask.
Focusing on the 1-shot case, where $K=1$, it begins with a feature extraction backbone network $f_B(\cdot)$, which encodes both $x^t$ and $x^r$ into semantic features $F^t$ and $F^r$ in $\mathbb{R}^{hw \times c}$, where $h$ and $w$ denote the height and width of the feature maps, and $c$ is the feature dimension.
Subsequent few-shot processes utilize these feature maps to generate a predicted segmentation $\tilde{y}\in \mathbb{R}^{H\times W}$ for $x^t$.
This prediction is then compared to the Ground Truth~(GT) $y^t$ for evaluation. 

The Segment Anything Model~(SAM) is a generalized foundation segmentation model adept at generating precise masks based on varied prompts of points, boxes, and coarse masks. 
Built around a core architecture that includes an image encoder, a prompt encoder, and a mask decoder, SAM effectively processes input images $x^t$ and prompts $P$ to produce detailed segmentation masks $\hat{y}$. 
These masks accurately delineate specific objects or regions within the images, based on the guidance provided by the prompts. 

\section{Method}
\begin{figure}[t]
    \centering    
    \includegraphics[width=1.0\linewidth]{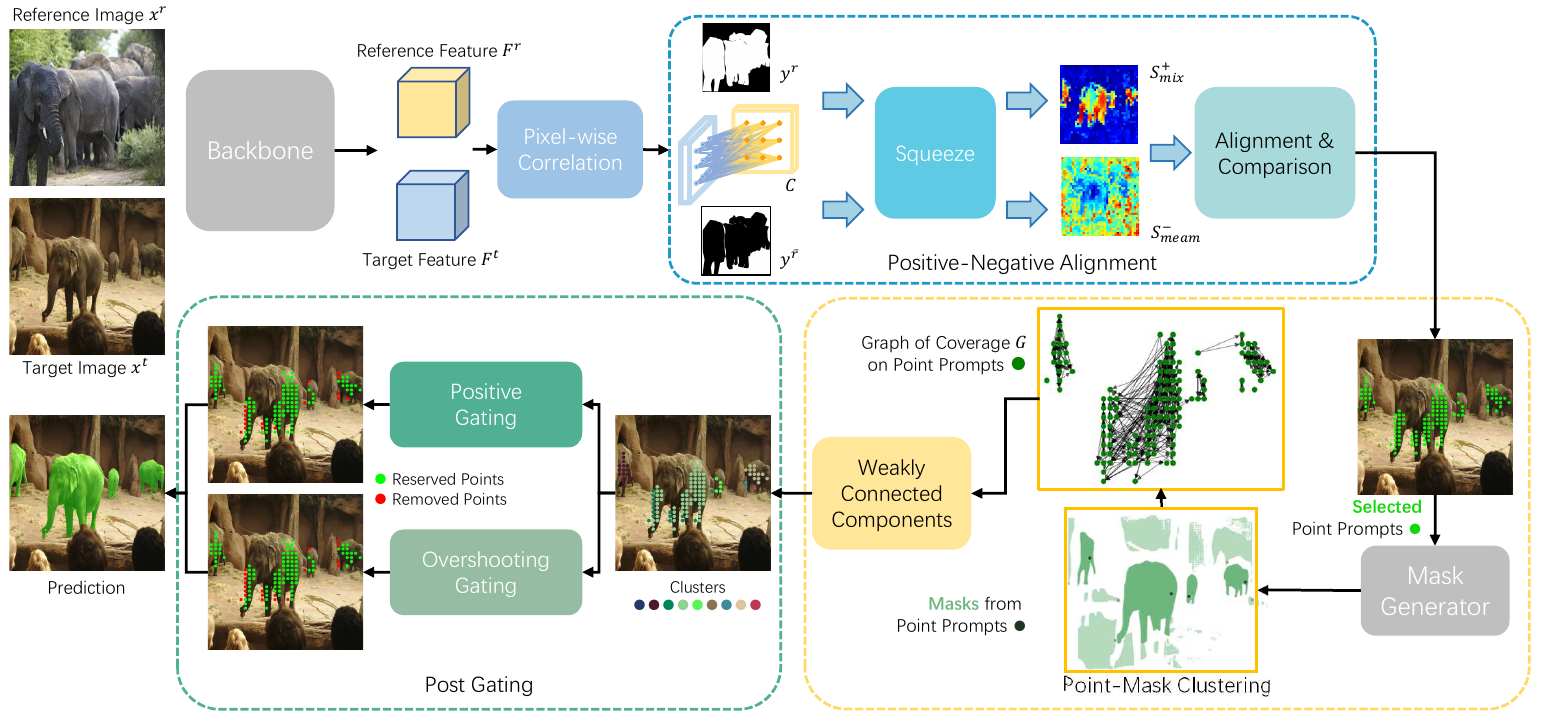}
    \caption{Overview of our approach, where the Positive-Negative Alignment module recognizes the correlation between target features and reference features for point selection, the Point-Mask Clustering module efficiently clusters the points based on the coverage of corresponding masks, and Post-Gating filters out the false-positive masks for generating final prediction.}
    \label{fig:framework}
\end{figure}

Diverging from traditional methods, we use a directed graph to exploit the natural relationships between points and their corresponding masks, representing fine-grained and coarse-grained features, respectively.
As shown in Fig.~\ref{fig:framework}, our approach mainly comprises the Positive-Negative Alignment~(PNA) module, Point-Mask Clustering~(PMC) module, and Post-Gating strategy. 
The PNA module leverages semantic features from the backbone network to sort pixel-wise correlations into similarity maps, enabling precise point selection.
The PMC module then clusters masks based on these selected points, while Post-Gating strategy refines the selection, enhancing the accuracy and reliability of the final prediction. 

\subsection{Positive-Negative Alignment for Point Selection}\label{sec:pna}

The PNA module efficiently selects point prompts to balance the number of points and coverage of target objects. 
Using the semantic features $F^r$ and $F^t$ from the reference and target images respectively~(with \textit{e.g.}, DINOv2~\cite{dinov223}), we get the pixel-wise correlation matrix $C \in \mathbb{R}^{hw \times hw}$: 
\begin{equation}
    \label{eq: cos}
    C(i, j) = ReLU\left( \frac{F^t(i)\cdot F^r(j)}{\|F^t(i)\|\cdot \|F^r(j)\|}\right), 
\end{equation}
where $C(i, j)$ represents the similarity between the $i$-th pixel of target features $F^t(i)$ and the $j$-th pixel of reference features $F^r(j)$. 

To minimize hyperparameter reliance, we leverage background features typically overlooked in FSS, indicated by the negative mask $y^{\tilde{r}}=\neg y^r$ of the reference image.
According to $y^r$ and $y^{\tilde{r}}$, we divide $C$ into $C^+$ and $C^-$ in $\mathbb{R}^{hw \times hw}$ for foreground and background features, respectively.
We then introduce two positive similarity maps in mean and max aspects respectively:
\begin{equation}\label{equ:2}
        S^+_{mean}(i) = \frac{\sum_{j=1}^{hw} C^+(i, j) }{\sum_{j=1}^{hw} \mathcal{I}(y^r)_j},\quad 
        S^+_{max}(i) = max(C^+(i)), 
\end{equation}
where $\mathcal{I}$ resizes $y^r$ to the same resolution as $F^r$ and then flatten it into a vector, $max(\cdot)$ finds the maximum value in the $i$-th row of $C^+$.
The mean positive similarity map $S^+_{mean} \in \mathbb{R}^{hw}$ captures global similarity towards the reference object but may blur distinct internal features, reducing accuracy for complex objects. 
In contrast, the max positive similarity map $S^+_{max} \in \mathbb{R}^{hw}$ focuses on the most similar regions, enhancing recall but also increasing noise. 
To maintain distinctiveness while reducing noise, we introduce the mixture similarity map $S^+_{mix} = S^+_{mean}\odot S^+_{max}$ using the Hadamard product. 
This method boosts target region distinctiveness by merging the strengths of both maps, while diminishing noise through the more stable global similarity. 

To select prompt points, we also use the mean negative similarity map $S^-_{mean}$, which reflects background similarity, noting that similar objects typically share higher background similarity values. 
We then align $S^+_{mix}$ and $S^-_{mean}$ by min-max normalization $\mathcal{M}$ to get: 
\begin{equation}
    S_p(i) = \mathcal{M}(S^+_{mix})(i) \cdot \mathbf{1}_{\{\mathcal{M}(S^+_{mix})(i) > \mathcal{M}(S^-_{mean})(i)\}},
\end{equation}
where $S_p\in \mathbb{R}^{hw}$ is the filtered map for point selection, and $\mathbf{1}_{\{\cdot\}}$ is 1 if the condition is true and 0 otherwise. 
Although we minimize false negatives, noise points remain.
To select suitable points from $S_p$ without hyperparameters, we define the sum of elements in $S_p$ as the number $N$ of points to be selected.
We then pick the $N$ highest-value points from $S_p$ as the point prompt set $\bm{P} = \{P_l\}_{l=1}^{N}$. 

\subsection{Point-Mask Clustering with Graph Connectivity}

We utilize point prompts from $\bm{P}$ to generate masks with SAM.
Each point $P_l$ in $\bm{P}$ corresponds to a unique mask $\hat{y}_l \in \mathbb{R}^{H\times W}$. 
As our point selection strategy prioritizes the coverage of objects, false-negative masks are unavoidable. 
Moreover, mask coverage can vary significantly within the same region, ranging from partial to full object coverage.
This necessitates understanding the internal relationships among coarse-grained masks and points from fine-grained feature comparison to ensure those covering the same target are accurately gathered.

To address this, we design the Point-Mask Clustering~(PMC) module, which clusters points and their corresponding masks based on mask coverage over points.
Following the principles of efficiency and automation, the PMC module is based on a directed graph $G = (V, E)$ with the vertex $v_l$ in $V$ representing point $P_l$ and its corresponding mask $\hat{y}_l$. 
Edges in $E$ are established based on mask coverage over other points; 
an edge $e_{l, m}$ exists if mask $\hat{y}_l$ covers points $P_m$ (with $m \neq l $).
Specifically, we do not establish edges for masks covering their corresponding points to avoid creating loops.

The graph $G$ is a directed simple graph, allowing us to cluster vertices by identifying weakly connected components.
This clustering process is hyperparameter-free, ensuring that every pair of vertices $u, v \in V$ within the same component has a directed path between them.
Each weakly connected component encompasses a set of points $\hat{P}_{p}$~(with $P=\{\hat{P}_{p}\}$) that are all covered by the union of their masks in $\hat{M}_{p}$, where $p$ indexes the clusters. 

The advanced SAM plays a crucial role in maintaining the precision of the generated masks. 
The precision of high-quality masks typically ensures non-overlapping between masks and prompting points of adjacent regions, especially those of different categories. 
This is the precondition for the efficacy of our PMC module, as even slight errors could significantly impact the clustering accuracy. 

\subsection{Post-Gating on Weakly Connected Components}

Our PNA module, while efficient in selecting points, inadvertently includes false positives, as detailed in Sec.~\ref{sec:pna}. 
To mitigate this, we have developed two gating strategies targeting distinct types of false positives based on clusters formed from weakly connected components.

\paragraph{Positive Gating.}
\label{sec:posjudge}

Despite the method in Sec.~\ref{sec:pna} diminishing the noise points outside the target region, there are still a few remaining noise points. 
These issues may have minimal impact on traditional segmentation methods, but under the SAM framework, masks derived from these noise points can significantly degrade accuracy. 
Moreover, some clusters of masks may extend beyond their intended target regions due to inaccuracies in SAM-generated masks or because the targeted object is part of a larger entity. 
Thus, we propose a Positive Gating strategy to address these issues. 

This strategy prioritizes mask effectiveness by assessing whether a mask contains more positive than negative pixels, thereby facilitating a specialized designed mask growth for final prediction.
The focus of mask growth is to enhance coverage of the target area rather than multiple objects, while minimizing background inclusion.
Firstly, this method utilizes a parameter-free gating mechanism that discriminates between pixel polarities, based on the positive and negative similarity maps, $S^+_{mean}$ and $S^-_{mean}$, as described in Sec.~\ref{sec:pna}.
To achieve this, we utilize $S^+_{mean}$ and $S^-_{mean}$, along with the median of $S^+_{mean}$~(\textit{i.e.}, the midpoint between the maximum and minimum values of $S^+_{mean}$), to constructs the polarity map $\bar{R}$ as follows:
\begin{equation}
    \bar{R}(i) = 
    \begin{cases}
        1, &  S^+_{mean}(i) \times S^+_{mean}(i) > s_{mid} \times S^-_{mean}(i), \\
        -1, & else. 
    \end{cases}\label{equ:4}
\end{equation}
Then, using the polarity map $\bar{R}$, we calculate the number of positive pixels of the $l^{th}$ mask as follows: 
\begin{equation}
    s^+_l = \sum_{i=1}^{hw} \bar{R}(i) \odot \mathcal{I}(\hat{y}_l)(i). 
\end{equation}
where $\mathcal{I}$ resizes and flattens $\hat{y}_l$ to the feature map's dimensions.
Subsequently, for each cluster $\hat{M}_{p}$ of weakly connected components, we sort the masks according to the ratio of positive pixel numbers to their areas. 
The indices of these sorted masks are denoted by $Q$. 
We then initialize a blank pseudo mask $\ddot{y}_p \in \mathbb{R}^{H\times W}$ and a set of positive points $P^+$.
Following this, we apply a Mask Growth algorithm as outlined in Sec.~\ref{sec:alg} and Alg.~\ref{algo:pj} for maintaining positive masks. 
This algorithm iteratively evaluates whether the region of $\hat{y}_q$ outside the pseudo mask $\ddot{y}_p$ is positive, updates $\ddot{y}_p$ with the identified positive mask, and adds its corresponding point into $P^+$.
    
\paragraph{Overshooting Gating.}

The fine-grained semantic features from $f(\cdot)$ are reliable for locating target objects, yet the point coverage of the target areas varies, leading to both under-coverage and over-coverage. 
SAM effectively addresses under-coverage; however, over-coverage, which extends beyond target boundaries, often produces false-positive masks.
These overshooting points, while semantically similar to the target areas in $F^t$, typically derive masks that cover areas outside the target, resulting in a mismatch of representations between the granularity of points and masks.
Thus, these points cannot be clustered with points inside the target areas. 
\begin{figure}[t]
    \centering    \includegraphics[width=\linewidth]{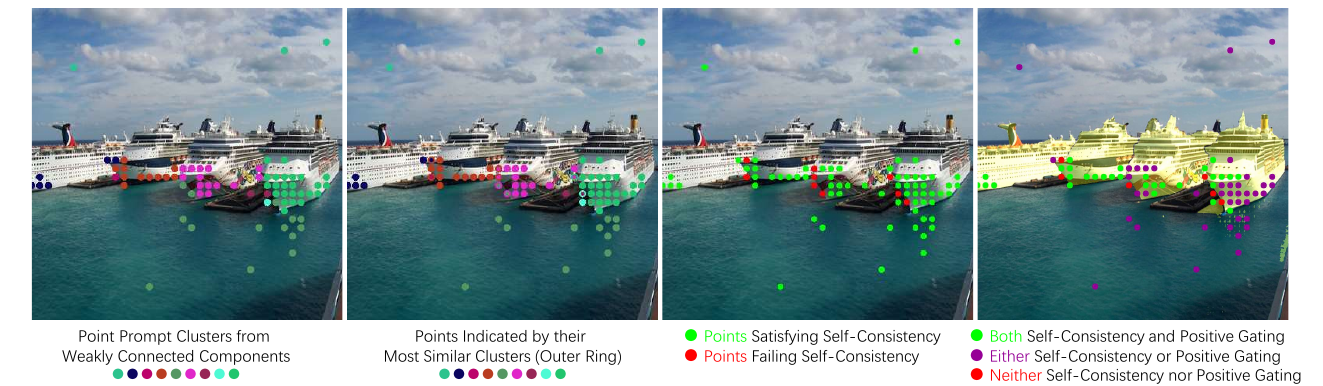}
    \caption{Illustration of the Overshooting Gating strategy.
    The outer ring of points in the second image indicates the most similar cluster of corresponding points, \textit{i.e.}, points with different outside and inside colors do not satisfy the self-consistency.}
    \label{fig:vissc}
\end{figure}

Hence, we devise an overshooting gating strategy with consideration of self-consistency to eliminate overshooting points and their associated masks. 
As shown in Fig.~\ref{fig:vissc}, We assess the similarity between the features of each point $P_l$ and the union mask $\hat{y}_{p} \in \mathbb{R}^{H\times W}$ from each mask cluster $\hat{M}_{p}$. 
The similarity computation for estimating \textbf{s}elf-\textbf{c}onsistency is performed as follows: 
\begin{equation}
    s^{sc}(l, p) = \frac{\sum_{i=1}^{hw} Sim(F^t(P_l), (F^t \odot \mathcal{I}(\hat{y}_{p}))(i))}{\sum_{i=1}^{hw} \mathcal{I}(\hat{y}_{p}) \cdot dist(l, p)}, 
\end{equation}
where $Sim(\cdot, \cdot)$ refers to the correlation calculation mentioned in Eq.~\ref{eq: cos}. 
We introduce an external function $dist(\cdot, \cdot)$ to measure the distance in $F^t$ between each point $P_l$ and the nearest selected point in $\hat{P}_{p}$. 
This measure helps confine comparison to neighboring clusters, minimizing interference from other instances. 
We then identify the cluster most similar to the points and retain those in the point set $P^{sc}$ that are more similar to their respective clusters. 

\paragraph{Mask Merging.}

Finally, we obtain two distinct sets of points, namely $P^+$ and $P^{sc}$. 
We then union the masks corresponding to points that are common to both $P^+$ and $P^{sc}$. 
The merged masks form the final prediction, denoted as $\tilde{y}$. 

\begin{table}[!t]
    \centering
    \caption{Performance on Few-shot Semantic Segmentation datasets of Pascal-5\textsuperscript{i}, COCO-20\textsuperscript{i}, FSS-1000, and LVIS-92\textsuperscript{i}. \desalt{Gray} means the in-domain trained results. 
    The best results are shown in \textbf{bold}. }
    \label{tab:fssex}
    \adjustbox{width=\linewidth}{
    \begin{tabular}{r|cc|cc|cc|cc}
    \toprule
         \multirow{2}{*}{Methods} & \multicolumn{2}{c|}{Pascal-5\textsuperscript{i}} & \multicolumn{2}{c|}{COCO-20\textsuperscript{i}} & \multicolumn{2}{c|}{FSS-1000} & \multicolumn{2}{c}{LVIS-92\textsuperscript{i}}\\
         & 1-shot & 5-shot &  1-shot & 5-shot & 1-shot & 5-shot & 1-shot & 5-shot \\
         \midrule
         \textit{\small specialist model} & & & & & & & &  \\
         HSNet~\cite{hsnet21}\pub{CVPR21} & \desalt{66.2} & \desalt{70.4} & \desalt{41.2} & \desalt{49.5} & \desalt{86.5} & \desalt{88.5} & 17.4 & 22.9 \\
         VAT~\cite{vat22}\pub{ECCV22} & \desalt{67.9} & \desalt{72.0} & \desalt{41.3} & \desalt{47.9} & \desalt{90.3} & \desalt{90.8} & 18.5 & 22.7 \\
         HDMNet~\cite{hdmnet23}\pub{CVPR23} & \desalt{69.4} & \desalt{71.8} & \desalt{50.0} & \desalt{56.0} & - & - & - & - \\
         AMFormer~\cite{amformer23}\pub{NeurIPS23} & \desalt{70.7} & \desalt{73.6} & \desalt{51.0} & \desalt{57.3} & - & - & - & - \\
         \midrule
         \textit{\small generalist model} & & & & & & & &  \\
         PerSAM~\cite{persam24}\pub{ICLR24} & 43.1 & - & 23.0 & - & 71.2 & - & 11.5 & - \\
         PerSAM-F~\cite{persam24}\pub{ICLR24} & 48.5 & - & 23.5 & - & 75.6 & - & 12.3 & - \\
         Matcher~\cite{matcher24}\pub{ICLR24} & 68.1 & 74.0 & 52.7 & 60.7 & 87.0 & \textbf{89.6} & 33.0 & 40.0 \\
         VRP-SAM~\cite{sun2024vrp}\pub{CVPR24} & \desalt{71.9} & - & \desalt{53.9} & - & - & - & - & - \\
         Ours & \textbf{72.1} & \textbf{82.6} & \textbf{58.7} & \textbf{66.8} & \textbf{88.0} & 88.9 & \textbf{35.2} & \textbf{44.2} \\
    \bottomrule
         
    \end{tabular}
    }
\end{table}
\begin{table}[!t]
    \centering
    \caption{Performance on One-shot Part Segmentation datasets and Cross Domain Few-shot Semantic Segmentation datasets. The best results are shown in \textbf{bold}. }
    \label{tab:opscdex}
    \adjustbox{width=\linewidth}{
    \begin{tabular}{r|c|c|cc|cc|cc}
    \toprule
         \multirow{3}{*}{Methods} & \multicolumn{2}{c|}{One-shot Part Seg.} & \multicolumn{6}{c}{Cross Domain FSS} \\
         & PASCAL-Part & PACO-Part &  \multicolumn{2}{c|}{Deepglobe} & \multicolumn{2}{c|}{ISIC} & \multicolumn{2}{c}{iSAID-5\textsuperscript{i}} \\
         & 1-shot & 1-shot &  1-shot & 5-shot & 1-shot & 5-shot & 1-shot & 5-shot \\
         \midrule
         \textit{\small specialist model}  & & & & & & & &  \\
         HSNet~\cite{hsnet21}\pub{CVPR21} & 32.4 & 22.6 & 29.7 & 35.1 & 31.2 & 35.1 & 34.1 & 40.4 \\
         DRA~\cite{dra24}\pub{CVPR24} & - & - & 41.3 & 50.1 & 40.8 & 48.9 & - & - \\
         FRINet~\cite{frinet23}\pub{TGRS23} & - & - & - & - & - & - & 42.6 & 44.5 \\
         \midrule
         \textit{\small generalist model}  & & & & & & & &  \\
         PerSAM~\cite{persam24}\pub{ICLR24} & 32.5 & 22.5 & 31.4 & - & 23.9 & - & 19.2 & - \\
         PerSAM-F~\cite{persam24}\pub{ICLR24} & 32.9 & 22.7 & 35.0  & - & 23.6 & - & 20.3 & - \\
         Matcher~\cite{matcher24}\pub{ICLR24} & 42.9 & 34.7 & 48.1 & 50.9 & 38.6 & 35.0 & 33.3 & 34.3 \\
         Ours & \textbf{44.5} & \textbf{36.3} & \textbf{49.5} & \textbf{57.7} & \textbf{48.7} & \textbf{55.2} & \textbf{47.1} & \textbf{52.4} \\
    \bottomrule         
    \end{tabular}}
\end{table}

\section{Experimental Results}


\subsection{Datasets}

To illustrate the Few-shot Semantic Segmentation ability and generalization capacity, we conduct three types of sub-tasks, \textit{i.e.} standard Few-shot Semantic Segmentation, One-shot Part Segmentation, and Cross Domain Few-shot Semantic Segmentation. 
The datasets for these tasks are as follows: 

\textbf{Pascal-5\textsuperscript{i}}, \textbf{COCO-20\textsuperscript{i}}, \textbf{FSS-1000}, and \textbf{LVIS-92\textsuperscript{i}} are standard FSS datasets. 
Pascal-5\textsuperscript{i}~\cite{ShabanOne} is based on the Pascal VOC 2012~\cite{pascal10} and SDS~\cite{sds11}. 
The 20 classes are separated into 4 folds of 5 classes. 
COCO-20\textsuperscript{i}~\cite{fwb19} is an 80-class dataset from MSCOCO~\cite{coco14}, which has 4 folds with each fold containing 20 classes. 
FSS-1000~\cite{fss1000} contains 1000 classes. 
The training, validation, and testing folds contain 520, 240, and 240 classes, respectively. 
LVIS-92\textsuperscript{i}~\cite{matcher24} is more challenging for evaluating generalist models, which select 920 classes with more than 2 images and divide these classes into 10 folds. 

\textbf{PASCAL-Part} and \textbf{PACO-Part}~\cite{matcher24} are One-shot Part Segmentation datasets. 
PASCAL-Part~\cite{pascalpart14,pascalpart20} contains 56 different object parts in 4 superclasses. PACO-Part is built based on the PACO dataset~\cite{paco23}, which has 456 object part classes. The 303 classes with at least 2 samples in PACO-Part are divided into four folds following Matcher~\cite{matcher24}. 

\textbf{Deepglobe}, \textbf{ISIC2018}, and \textbf{iSAID-5\textsuperscript{i}} are Cross Domain FSS datasets. 
The Deepglobe~\cite{deepglobe} contains satellite images of geographic categories including urban, agriculture, rangeland, forest, water, and barren. 
The ISIC2018~\cite{isic18} is a skin lesion analysis dataset with three classes. 
The iSAID-5\textsuperscript{i}~\cite{isaid521} evenly split 3 folds for 15 classes based on the remote sensing dataset iSAID~\cite{isaid19}. 



\subsection{Implementation Details}

Following the settings of PerSAM~\cite{persam24} and Matcher~\cite{matcher24} for a fair comparison, we use DINOv2~\cite{dinov223} with a ViT-L/14~\cite{vit20} as our feature extraction backbone, and SAM~\cite{sam23} with ViT-H as the mask generator. 
The input image sizes are set to $518\times 518$ for DINOv2 and $1024\times 1024$ for SAM following Matcher~\cite{matcher24}. 
Except for the default hyperparameters of SAM and DINOv2, our approach \textbf{do not have any external hyperparameter}. 
We apply the mean Intersection over Union~(mIoU) metric for evaluating the performance. 
All experiments are conducted on a single NVIDIA RTX2080Ti. 

\subsection{Comparison with State-Of-The-Arts}

\textbf{Comparison on the standard FSS datasets. } 
We compared our approach with other state-of-the-art specialist and generalist models. 
As shown in Tab.~\ref{tab:fssex}, our approach achieves 72.1\% mIoU on the Pascal-5\textsuperscript{i} dataset and 58.7\% mIoU on COCO-20\textsuperscript{i} dataset, which surpasses all previous specialist and generalist state-of-the-art models. 
Our approach reaches 35.2\% mIoU in the more challenging dataset of LVIS-92\textsuperscript{i}, with 2.2\% of improvement compared to the previous training-free method Matcher. 
The performance remains competitive on the FSS-1000 compared with specialist models. 
The 5-shot result of Pascal-5\textsuperscript{i}, COCO-20\textsuperscript{i}, and LVIS-92\textsuperscript{i} further extends the lead, which proves that our approach can effectively handle the few-shot scenario. 

\textbf{Comparison on the One-shot Part Segmentation datasets. }
The One-shot Part Segmentation tasks evaluate the ability to fetch the target part from the whole object. 
The results in Tab.~\ref{tab:opscdex} show that our approach achieves the mIoU of 44.5\% and 36.3\% on both datasets of PASCAL-Part and PACO-Part, respectively. 
Our approach outperforms the state-of-the-art generalist model Matcher with 1.6\% on both datasets. 
Given that Matcher employs \textbf{specific hyperparameters} to enhance part segmentation, our superior performance demonstrates the adaptability of our approach across both object and part segmentation contexts. 

\textbf{Comparison on the Cross Domain FSS datasets. }
The Cross Domain FSS tasks validate the performance on different domains. 
Our approach achieves state-of-the-art performance in datasets of Deepglobe, ISIC, and iSAID-5\textsuperscript{i} among other specialist domain models and generalist models. 
Especially within the context of the skin lesion analysis dataset ISIC and remote sensing dataset iSAID-5\textsuperscript{i}, our approach outperforms Matcher by margins of 10.1\% and 13.8\% respectively.

\subsection{Ablation Study}

\begin{table*}[t] 
    \centering
    \begin{minipage}{0.46\linewidth} 
        \centering
        \caption{Ablation study of Point Selection.}
        \label{tab:abps}
        \adjustbox{width=\linewidth}{
        \begin{tabular}{ccc|c|c}
            \toprule
            $S^+_{mean}$ & $S^+_{max}$ & $S^-_{mean}$ &Top $N$ & mIoU\\
            \midrule
            \checkmark & & \checkmark & \checkmark & 53.1 \\
             & \checkmark & \checkmark & \checkmark & 54.1 \\
            \checkmark & \checkmark & \checkmark & & 56.4 \\
            \checkmark & \checkmark & & \checkmark & 51.5 \\
            \checkmark & \checkmark & \checkmark & \checkmark & \textbf{58.7}\\
            \bottomrule
        \end{tabular}}
    \end{minipage}
    \hfill 
    \begin{minipage}{0.49\linewidth} 
        \centering
        \caption{Ablation study of PMC and Post-Gating. }
        \label{tab:abj}
        \adjustbox{width=\linewidth}{
        \begin{tabular}{cc|cc|cc}
            \toprule
            \multicolumn{2}{c|}{PG} & \multicolumn{2}{c|}{OG} & COCO-20\textsuperscript{i} & LVIS-92\textsuperscript{i} \\
            Strong & Weak & Strong & Weak & & \\
            \midrule
             & & & & 44.0 & 24.2 \\
              & \checkmark & & & 57.1 & 34.3 \\
            \checkmark & & & & 57.1 & 33.9 \\
            & \checkmark & \checkmark & & 56.7 & \textbf{35.2} \\
             & \checkmark & & \checkmark & \textbf{58.7} & \textbf{35.2} \\
            \midrule
            \multicolumn{4}{c|}{k-means++} & \desalt{57.5} & \desalt{34.0} \\
            \bottomrule
        \end{tabular}}
    \end{minipage}
\end{table*}

\begin{table*}[t] 
    \centering
    \begin{minipage}{0.52\linewidth} 
        \centering
        \caption{Ablation study of positive gating on each cluster. M.G. represents the Mask Growth algorithm. }
        \label{tab:abpj}
        \adjustbox{width=\linewidth}{
        \begin{tabular}{cc|cc}
            \toprule
            Strategies & M.G. & COCO-20\textsuperscript{i} &  PASCAL-Part \\
            \midrule
            \multirow{2}{*}{Sum} & & 55.3 & 39.1 \\
             & \checkmark & 58.6 & 44.3 \\
             \midrule
            \multirow{2}{*}{Num} & & 57.1 & 42.2 \\
            & \checkmark & \textbf{58.7} & \textbf{44.5}\\
            \bottomrule
        \end{tabular}}
    \end{minipage}
    \hfill 
    \begin{minipage}{0.42\linewidth} 
        \centering
        \caption{Ablation on the strategies of Self-Consistency measurement. }
        \label{tab:absc}
        \adjustbox{width=0.9\linewidth}{
        \begin{tabular}{r|cc}
            \toprule
            Strategies & mIoU & $\Delta$ \\
            \midrule
            None & 57.1 & 0.0 \\
            Point Sim. & 56.7 & -0.4 \\
            MAP Sim. & 57.7 & +0.6 \\
            Mean Sim. W/o dist & 49.1 & -8.0 \\
            Mean Sim. (Ours) & \textbf{58.7} & \textbf{+1.6} \\
            \bottomrule
        \end{tabular}}
    \end{minipage}
\end{table*}


\textbf{Point Selection.} We evaluate the impact of various similarity maps and the parameter-free selection of top N points on performance, as detailed in Sec.~\ref{sec:pna}. 
As shown in Tab.~\ref{tab:abps}, using either $S^+_{mean}$ or $S^+_{max}$ alone leads to a performance drop of up to $5.6\%$ compared to using both. 
This decline is due to the inherent limitations of $S^+_{mean}$ and $S^+_{max}$ discussed in Sec~\ref{sec:pna}. 
Additionally, the evaluation confirms that picking the top-N points based on similarity, which is parameter-free and requires no additional settings, simplifies the process and increases accuracy by $2.1\%$.


\textbf{Clustering Method.} We compare our PMC module using weakly connected components with the PMC module using strong connected components, which provides finer clustering results. 
According to our experiment results in Tab.~\ref{tab:abj}, the clusters from weakly connected components provide better performance on COCO-20\textsuperscript{i} and LVIS-92\textsuperscript{i} for both gating, as these clusters of masks have ideal coverage of the objects. 
Simply filtering the masks without clustering-based gating can only achieve 44.0\% mIoU on COCO-20\textsuperscript{i} and 24.2\% mIoU on LVIS-92\textsuperscript{i}, which is significantly lower than the performance achieved with clustering-based gating. 
Furthermore, our dynamic hyperparameter-free clustering method outperforms the k-means++ with 1.2\% on both datasets. Note that k of k-means++ is set to 10 following Matcher~\cite{matcher24}. 

\begin{figure}[!tb]
    \centering    \includegraphics[width=1.0\linewidth]{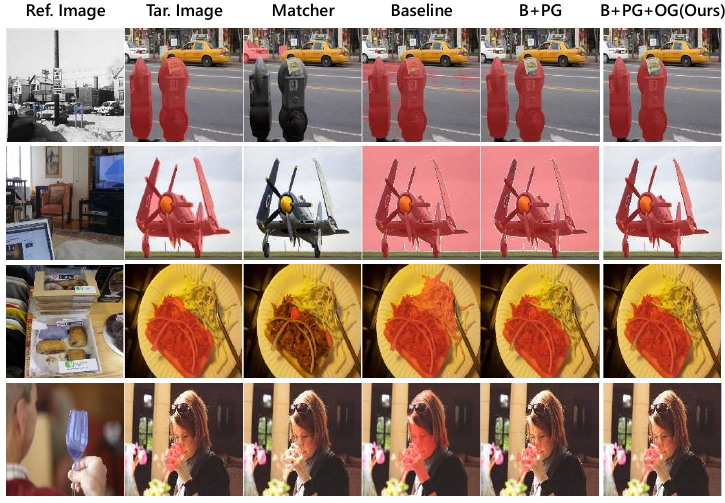}
    \caption{Qualitative analysis of Matcher, Baseline, B+PG, B+PG+OG. B, PG, and OG respectively represent Baseline, Positive Gating, and Overshooting Gating. Masks in ref. image are shown in \textcolor{RoyalBlue}{blue}.}
    \label{fig:qualitative}
\end{figure}

\textbf{Positive Gating.} Our approach compares the number of positive points and negative points in $\hat{S}^+$~(Num) to judge whether the mask is positive. 
We conduct experiments for the strategy of comparing the sum of positive and negative values~(Sum). 
The results in Tab.~\ref{tab:abpj} demonstrate the Num strategy yields better performance, as comparing the number of pixels mitigates the influence of a few excessively high similarity values. 
Furthermore, the utilization of the Mask Growth algorithm improves both FSS and Part Segmentation performance by carefully retaining the positive regions. 
However, it weakens the improvement of Num due to their similar effects. 


\textbf{Overshooting Gating.} Our Overshooting gating aims to filter out the overshooting points closely neighboring to target regions, thus having a less remarkable improvement of 1.6\% compared to Positive Gating, as shown in Tab.~\ref{tab:absc}. 
This performance still surpasses the mean similarity of comparing points with regions of clustered points~(Point Sim.) or the prototypes from Masked Average Pooling~\cite{sgone20} with union masks~(MAP Sim.). 
More importantly, the distance function avoids the gating from 9.6\% of performance decline. It ensures each cluster only affects neighboring points. 

\subsection{Qualitative Analysis}
\label{sec:qual}

Here we present the qualitative results of Matcher, Baseline~(1\textsuperscript{st} row in Tab.~\ref{tab:abj}), Baseline+PG~(2\textsuperscript{nd} row in Tab.~\ref{tab:abj}) and our approach in Fig.~\ref{fig:qualitative}. 
The bipartite matching of Matcher has a negative influence when the areas of the target object in reference and target images have significant differences, as shown in the 1\textsuperscript{st} and 3\textsuperscript{rd} rows. 
The positive gating with clustering filters out the noise masks in the 3\textsuperscript{rd} row, while the Overshooting Gating further removes the masks belonging to overshooting points in the 2\textsuperscript{nd} and 4\textsuperscript{th} rows. 
More qualitative analyses are conducted in the appendix. 

\section{Conclusions}

In this paper, we propose an efficient, training-free SAM-based FSS approach that requires no external hyperparameters. 
As an automatic SAM-based semantic segmentation pipeline, our approach balances candidate points and object coverage in the Positive-Negative Alignment~(PNA) module, then uses SAM-generated masks in the Point-Mask Clustering~(PMC) module to enhance Post Gating.
Extensive experiments demonstrate the superior performance of our approach, advancing semantic segmentation without extensive parameter tuning or training.

\textbf{Acknowledgment.} 
This work was supported by the National Natural Science Foundation of China under No. 62472033, No. U23A20314, and No. 61972036. J. Jiao is supported by the Royal Society Short Industry Fellowship (SIF\textbackslash R1\textbackslash231009) and the Amazon Research Award.

\bibliographystyle{ieeetr}
\bibliography{reference}

\clearpage
\appendix
\section{Appendix / supplemental material}

\subsection{More Details for Mask Growth Algorithm}
\label{sec:alg}

We mention the Mask Growth algorithm in Sec.~\ref{sec:posjudge}. 
The Mask Growth algorithm is designed for each cluster of masks $\hat{M}_{weak, p}$. 
The details of the algorithm are shown in Alg.~\ref{algo:pj}. 
We first initialize an empty set $P^+$ and a blank pseudo mask $\ddot{y}_p$. 
Then, we start an iterative process and get the current mask $\hat{y}_q$ based on the sorted sequence of indices $Q$. 
The parts of the current mask $\hat{y}_q$ overlapping with $\ddot{y}_p$ are removed. 
We compute the positive value $s_q^+$ of the remaining parts. 
If $s_q^+$ is positive, the mask $\hat{y}_q$ is updated into the $\ddot{y}_p$ and its corresponding point $P_q$ is added into $P^+$. 
As soon as the iterative process finishes, the set of positive points $P^+$ is established. 

\begin{algorithm}
    \renewcommand{\algorithmicrequire}{\textbf{Input:}}
    \renewcommand{\algorithmicensure}{\textbf{Output:}}
    \caption{Mask Growth for each cluster}
    \label{algo:pj}
    \begin{algorithmic}
        \REQUIRE $\hat{M}_{p}, \ddot{y}_p, Q$, $P^+$
        \FOR{$n=1$ to $|Q|$}
        \STATE $q \gets Q(n)$
        \STATE $\hat{y}_q \gets \hat{M}_p(q) $
        \STATE $\hat{y}_q = \hat{y}_q \& \sim \ddot{y}_p$ 
        \STATE $s^+_q \gets \sum_{i=1}^{hw} \hat{S}^+(i) \odot \mathcal{I}(\hat{y}_q)(i) $
        \IF{$s^+_q > 0$}
            \STATE $\text{Add}~P_q~\text{to}~P^+. $
            \STATE $\ddot{y}_p = \hat{y}_q \vee \ddot{y}_p$
        \ENDIF
        \ENDFOR
        \ENSURE $P^+$
    \end{algorithmic}
\end{algorithm}

\subsection{Limitations}

Our approach has impressive performance on Few-shot Semantic Segmentation tasks. 
However, due to the resolution of features $F^t$ from DINOv2 not aligning with the required resolution for prompting the SAM, we directly map the coordinates of points in $F^t$ to coordinates for prompting. 
This results in coordinate bias for small objects, as the gap between neighboring points can reach approximately 28 pixels. 
Our future work will focus on locating small objects. 

\subsection{Societal Impacts}

As a completely automatic SAM-based few-shot semantic segmentation approach without external hyperparameters, our method is capable of handling various scenarios of semantic segmentation, as demonstrated by our extensive experiments. 
The efficiency and generalizability of our method ensure a wide range of applications. 
Furthermore, since our training-free method is constructed upon the widely used open-source foundation models, we have not identified the negative societal impact to date.

\subsection{Details of Current SAM-based FSS Methods}

\begin{figure}[t]
    \centering
    \subfigure[]{
        \label{fig:persam}        \includegraphics[width=0.18\linewidth]{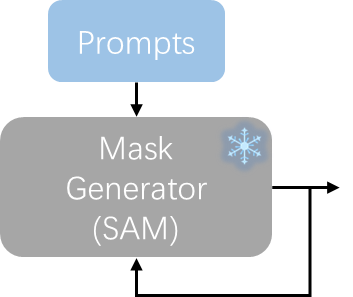}
    }
    \hfill
    \subfigure[]{
        \label{fig:matcher}        \includegraphics[width=0.42\linewidth]{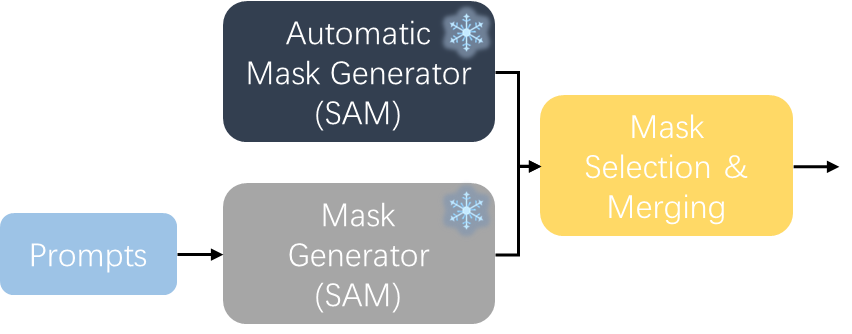}
    }
    \hfill
    \subfigure[]{
        \label{fig:ours_simple}        \includegraphics[width=0.32\linewidth]{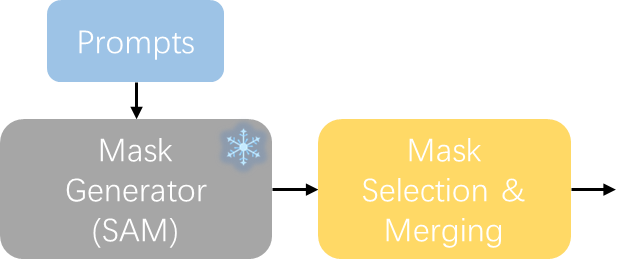}
    }
    \caption{Comparison of the pipeline between the previous methods and our approach. (a) PerSAM~\cite{persam24} iteratively uses the Mask Generator to refine the mask. (b) Matcher~\cite{matcher24} introduced an external Automatic Mask Generator~\cite{sam23} with automatic prompting to excessively generate masks from the whole image. (c) The effectiveness of the PMC module and Post-Gating ensures that our approach uses Mask Generator with our prompts only once. }
    \label{fig:simple}
\end{figure}

Our approach aims to address several issues present in previous SAM-based FSS methods to achieve an automatic SAM-based model. 
These issues include the requirement of excessive external hyperparameters, overusing the mask generator of SAM, prolonged inference times, etc. 
The Tab.~\ref{tab:set_samfss} shows the difference between our approach and previous SAM-based FSS methods. 
Fig.~\ref{fig:simple} shows the difference in using SAM as the mask generator between our approach and previous methods. 
Fig.~\ref{fig:persam} presents the iterative refinement of PerSAM, which involves generating masks from SAM 3 times. 
Fig.~\ref{fig:matcher} exhibits that Matcher introduces an external Automatic Mask Generator, which automatically prompts for generating all mask proposals in the image. 
Our approach in Fig.~\ref{fig:ours_simple} only utilizes the standard Mask Generator of SAM and generates the masks with our prompts only once. 

\begin{table}[h]
    \caption{Details of the current SAM-based FSS methods. }
\small
    \centering
    \begin{tabular}{r|ccccc}
    \toprule
         Methods & PerSAM & PerSAM-F & Matcher & VRP-SAM & Ours \\
         \midrule
         Training-free & \checkmark & & \checkmark & & \checkmark \\
         External-hyperparameters-free & \checkmark & & & & \checkmark \\
         Once mask generation & & & & \checkmark & \checkmark \\
         Inference speed~(s/img) & 1.43 & 16.5 & 12.7 & N/A & 1.88 \\
         COCO-20\textsuperscript{i} mIoU & 23.0 & 23.5 & 52.7 & 53.9 & 58.7\\
         \bottomrule
    \end{tabular}
    \label{tab:set_samfss}
\end{table}

\subsection{Simple Discussion of SAM}

\begin{figure}
    \centering
    \includegraphics[width=0.95\linewidth]{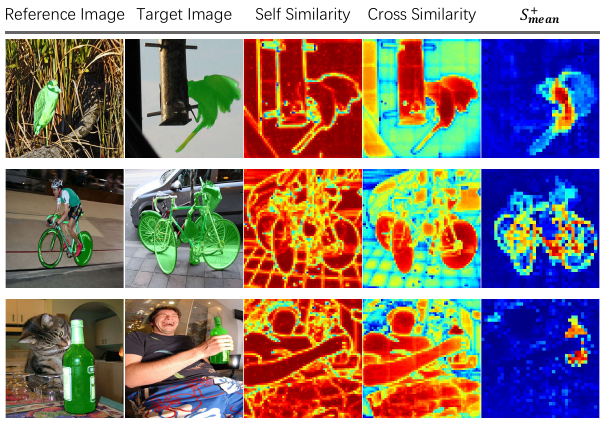}
    \caption{Analysis of the features from default ViT encoder of SAM. }
    \label{fig:sama}
\end{figure}

\subsubsection{Features from ViT Encoder of SAM}

Previous state-of-the-art generalist FSS methods~\cite{} use DINOv2 or ResNet-50, instead of the default ViT encoder of SAM, for fine-grained features. 
We visualize the representative samples of Pascal-5\textsuperscript{i} in Fig.~\ref{fig:sama}. 
The 3\textsuperscript{rd} column of maps represents the self-similarity of the $F^t_{SAM}$. We introduce the $3\times 3$ average pooling for $F^t_{SAM}$ followed by computing the cosine similarity between the pooled features and $F^t_{SAM}$. 
The maps illustrate that $F^t_{SAM}$ can accurately identify the regions of objects within the image, where the features within each object region are nearly identical, while features between neighboring different objects are distinct. 

Although the characteristics of $F^t_{SAM}$ ensure the generation of high-quality masks, the coarse-grained features are not suitable for locating the objects, as shown in the 4\textsuperscript{th} column. 
The similarity between $F^t_{SAM}$ and $F^r_{SAM}$ cannot effectively distinguish the target object well compared to $S^+_{mean}$ from DINOv2. 
Therefore, we follow the previous methods using DINOv2 for fine-grained features. 

\subsubsection{Masks analysis for Point-Mask Clustering}

\begin{figure}[!t]
    \centering
    \includegraphics[width=0.65\linewidth]{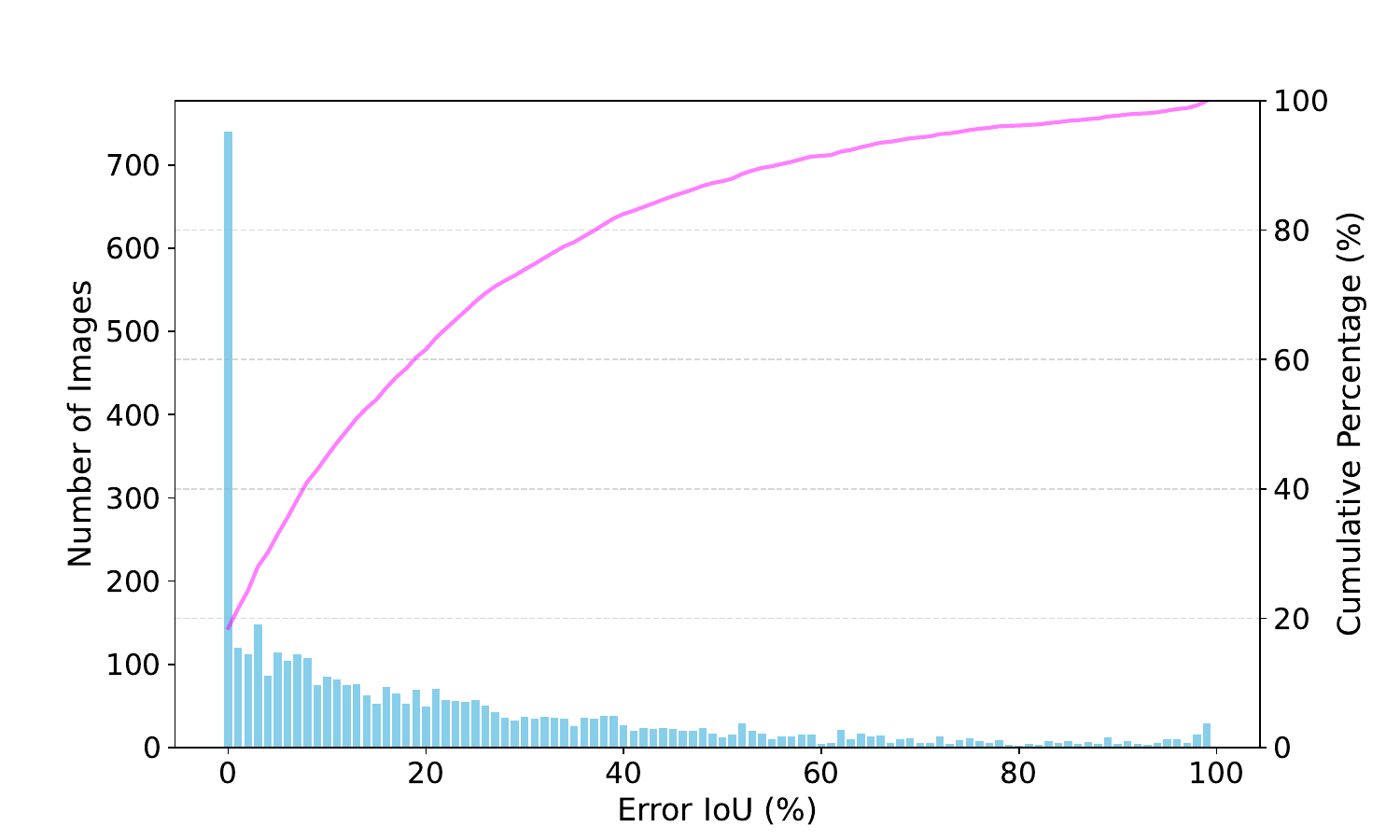}
    \caption{The distribution of IoU between masks from foreground points and from background points. }
    \label{fig:erroriou}
\end{figure}

\begin{figure}[!t]
    \centering
    \includegraphics[width=0.65\linewidth]{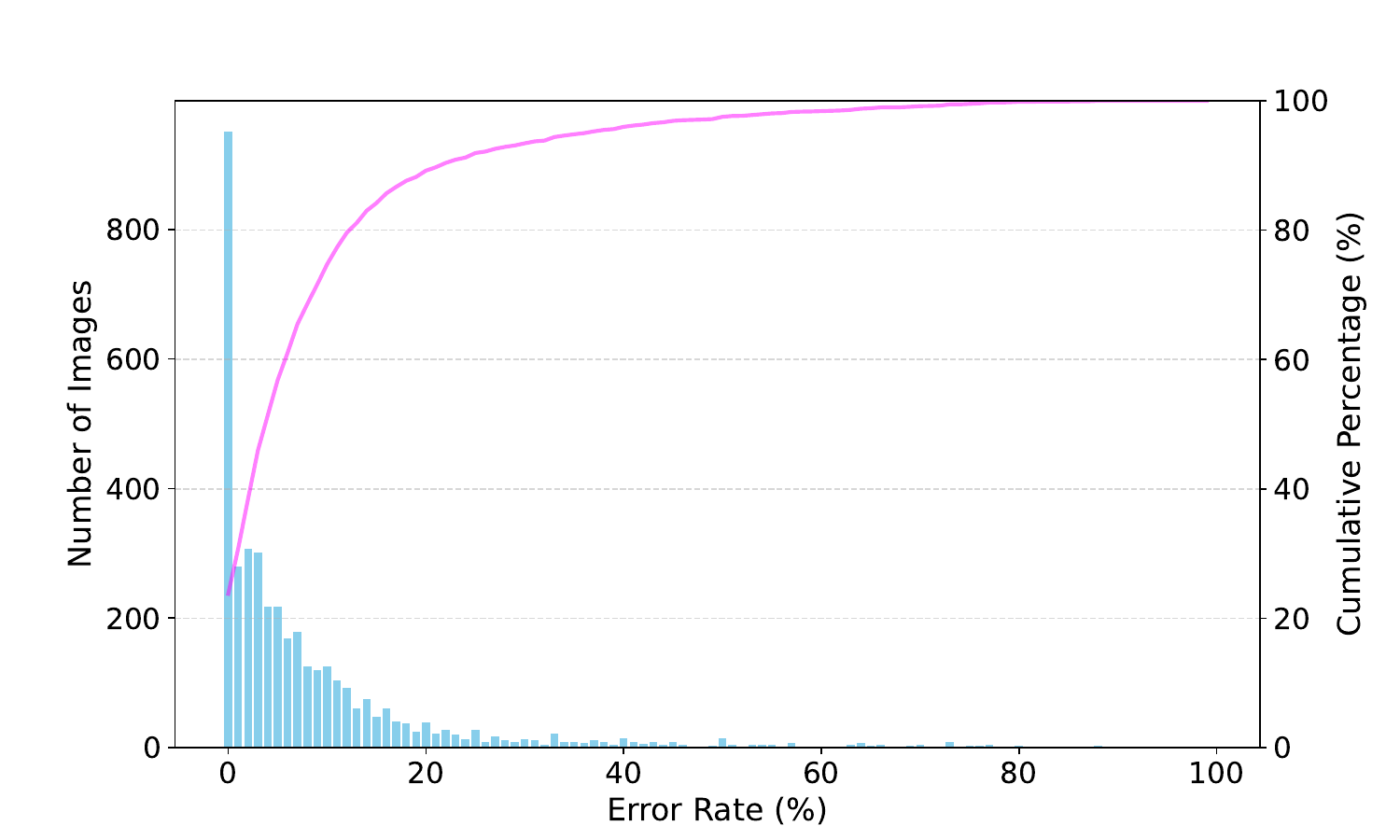}
    \caption{The distribution of the ratio between the number of background points and all points covered by the masks from foreground points. }
    \label{fig:errorrate}
\end{figure}

\begin{figure}[!t]
    \centering
    \includegraphics[width=0.65\linewidth]{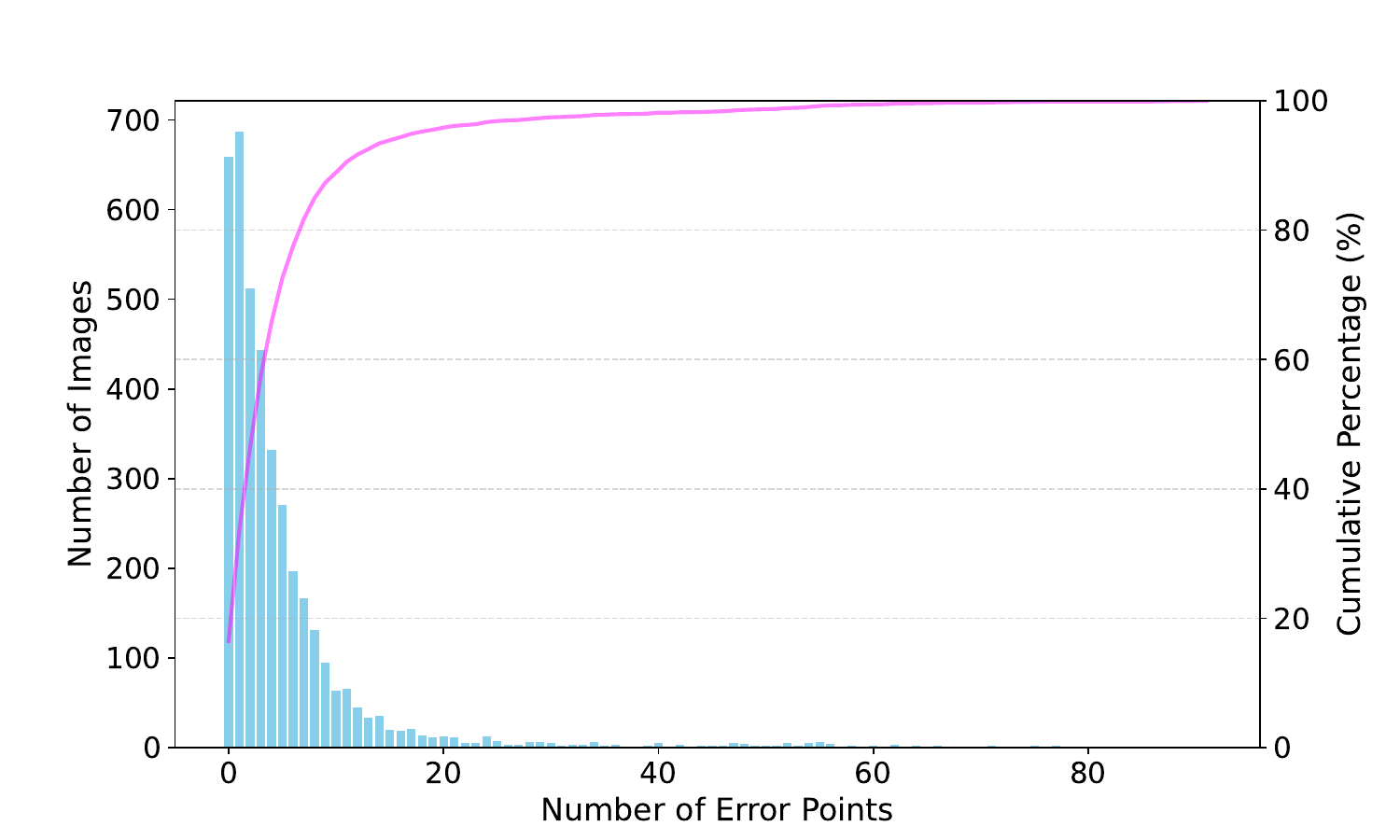}
    \caption{The distribution of the number of background points covered by the masks from foreground points. }
    \label{fig:errornum}
\end{figure}

Our Point-Mask Clustering module introduces a parameter-free clustering method by constructing a graph of coverage. 
The effectiveness of the method primarily relies on the high-quality masks, whose boundaries mostly align with the object boundaries. 
We roughly analyze the coverage of masks generated from the points in the ground truth foreground region using 4000 samples from Pascal-5\textsuperscript{i}. 
In particular, we get the union of masks from the foreground points as $\hat{y}_{fore}$, and visualize three distributions, including the distribution of IoU between $\hat{y}_{fore}$ and union of masks from background points $\hat{y}_{back}$ in Fig.~\ref{fig:erroriou}, the ratio between the number of background points and all points covered by the $\hat{y}_{fore}$ in Fig.~\ref{fig:errorrate}, the number of background points covered by $\hat{y}_{fore}$ in Fig.~\ref{fig:errornum}. 
The distribution charts demonstrate that most of the samples have acceptable coverage on the error points for Point-Mask Clustering. 
Given the limited precision of ground truth annotations, the analysis is for reference only. 
The effectiveness of our Point-Mask Clustering is validated in our ablation study in Tab.~\ref{tab:abj}.

\subsection{Additional Experiment Results. }

\subsubsection{Performance of Different Foundation Model Sizes}

Tab.~\ref{tab:size} shows the experiment results of our approach with different scales of SAM and DINOv2. 
Compared to the previous training-free method Matcher, our approach still achieves a better performance with SAM-Large and DINOv2-Base. 
The fair comparison with SAM-Huge and DINOv2-Large further demonstrates the effectiveness of our approach.

\begin{table}[h]
    \centering
    \caption{Evaluation of our approach with different sizes of SAM and DINOv2. }
    \begin{tabular}{c|cc|c|ccc}
    \toprule
         Methods & SAM & DINOv2 & Params. & COCO-20\textsuperscript{i} & FSS-1000 & LVIS-92\textsuperscript{i} \\
         \midrule
         Matcher & huge & large & 945M & 52.7 & 87.0 & 31.4 \\
         \midrule
         \multirow{4}{*}{Ours} & base & base & 180M & 53.7 & 85.6 & 31.1 \\
         & large & base & 399M & 55.5 & 87.5 & 31.7 \\
         & large & large & 617M & 58.0 & 87.8 & 35.1 \\
         & huge & large & 945M & 58.7 & 88.0 & 35.2 \\
         \bottomrule
    \end{tabular}
    \label{tab:size}
\end{table}

\subsubsection{Detailed Results of Evaluation Datasets}

We present the detailed results on different Few-shot Semantic Segmentation datasets, including Pascal-5\textsuperscript{i} in Tab.~\ref{tab:dpas}, COCO-20\textsuperscript{i} in Tab.~\ref{tab:dcoco}, LVIS-92\textsuperscript{i} in Tab.~\ref{tab:dlvis}, PASCAL-Part and PACO-Part in Tab.~\ref{tab:dpart}, iSAID-5\textsuperscript{i} in Tab.~\ref{tab:disaid}. 
The results show that our approach has remarkable performance in each fold of the datasets, demonstrating its generalized effectiveness in various scenarios.  

\begin{table}[!h]
    \centering
    \caption{Detail results of Pascal-5\textsuperscript{i}. }
    \begin{tabular}{c|ccccc|ccccc}
    \toprule
        \multirow{2}{*}{Methods} & \multicolumn{5}{c|}{Pascal-5\textsuperscript{i} 1-shot} & \multicolumn{5}{c}{Pascal-5\textsuperscript{i} 5-shot} \\
        & fold0 & fold1 & fold2 & fold3 & mean & fold0 & fold1 & fold2 & fold3 & mean \\
        \midrule
        AMFormer & 71.3 & 76.7 & 70.7 & 63.9 & 70.7 & 74.4 & 78.5 & 74.3 & 67.2 & 73.6 \\
        Matcher & 67.7 & 70.7 & 66.9 & 67.0 & 68.1 & 71.4 & 77.5 & 74.1 & 72.8 & 74.0 \\
        Ours & 71.1 & 75.7 & 69.2 & 73.3 & 72.1 & 81.5 & 86.3 & 79.7 & 82.9 & 82.6 \\
        \bottomrule
    \end{tabular}
    \label{tab:dpas}
\end{table}

\begin{table}[!h]
    \centering
    \caption{Detail results of COCO-20\textsuperscript{i}. }
    \begin{tabular}{c|ccccc|ccccc}
    \toprule
        \multirow{2}{*}{Methods} & \multicolumn{5}{c|}{COCO-20\textsuperscript{i} 1-shot} & \multicolumn{5}{c}{COCO-20\textsuperscript{i} 5-shot} \\
        & fold0 & fold1 & fold2 & fold3 & mean & fold0 & fold1 & fold2 & fold3 & mean \\
        \midrule
        AMFormer & 44.9 & 55.8 & 52.7 & 50.6 & 51.0 & 52.0 & 61.9 & 57.4 & 57.9 & 57.3 \\
        Matcher & 52.7 & 53.5 & 52.6 & 52.1 & 52.7 & 60.1 & 62.7 & 60.9 & 59.2 & 60.7 \\
        Ours & 56.6 & 61.4 & 59.6 & 57.1 & 58.7 & 67.1 & 69.4 & 66.0 & 64.8 & 66.8 \\
        \bottomrule
    \end{tabular}
    \label{tab:dcoco}
\end{table}

\begin{table}[!h]
    \centering
    \caption{Detail results of LVIS-92\textsuperscript{i}. }
    \begin{tabular}{c|ccccccccccc}
    \toprule
        \multirow{2}{*}{Methods} & \multicolumn{11}{c}{LVIS-92\textsuperscript{i} 1-shot} \\
        & fold0 & fold1 & fold2 & fold3 & fold4 & fold5 & fold6 & fold7 & fold8 & fold9 & mean \\
        \midrule
        Matcher & 31.4 & 30.9 & 33.7 & 38.1 & 30.5 & 32.5 & 35.9 & 34.2 & 33.0 & 29.7 & 31.4 \\
        Ours & 30.9 & 37.9 & 37.1 & 39.6 & 31.2 & 36.4 & 39.1 & 35.7 & 32.3 & 31.5 & 35.2 \\
        \midrule
        \multirow{2}{*}{Methods} & \multicolumn{11}{c}{LVIS-92\textsuperscript{i} 5-shot} \\
         & fold0 & fold1 & fold2 & fold3 & fold4 & fold5 & fold6 & fold7 & fold8 & fold9 & mean \\
         \midrule
        Matcher & 37.0 & 36.6 & 47.3 & 39.1 & 37.1 & 41.8 & 42.7 & 37.7 & 37.9 & 43.3 & 40.0 \\
        Ours & 42.1 & 38.4 & 50.0 & 42.5 & 42.0 & 46.5 & 46.4 & 41.5 & 43.7 & 48.4 & 44.2 \\
        \bottomrule
    \end{tabular}
    \label{tab:dlvis}
\end{table}

\begin{table}[!h]
    \centering
    \caption{Detail results of PASCAL-Part and PACO-Part. }
    \begin{tabular}{c|ccccc|ccccc}
    \toprule
        \multirow{2}{*}{Methods} & \multicolumn{5}{c|}{PASCAL-Part} & \multicolumn{5}{c}{PACO-Part} \\
        & animals & indoor & person & vehicles & mean & fold0 & fold1 & fold2 & fold3 & mean \\
        \midrule
        HSNet & 21.2 & 53.0 & 20.2 & 35.1 & 32.4 & 20.8 & 21.3 & 25.5 & 22.6 & 22.6 \\
        Matcher & 37.1 & 56.3 & 32.4 & 45.7 & 42.9 & 32.7 & 35.6 & 36.5 & 34.1 & 34.7 \\
        Ours & 33.2 & 59.6 & 35.2 & 50.1 & 44.5 & 33.4 & 34.9 & 39.7 & 37.0 & 36.3 \\
        \bottomrule
    \end{tabular}
    \label{tab:dpart}
\end{table}

\begin{table}[!h]
    \centering
    \caption{Detail results of iSAID-5\textsuperscript{i}. }
    \begin{tabular}{c|cccc|cccc}
    \toprule
        \multirow{2}{*}{Methods} & \multicolumn{4}{c|}{iSAID-5\textsuperscript{i} 1-shot} & \multicolumn{4}{c}{iSAID-5\textsuperscript{i} 5-shot} \\
        & fold0 & fold1 & fold2 & mean & fold0 & fold1 & fold2 & mean \\
        \midrule
        FRINet & 46.5 & 36.9 & 43.9 & 42.6 & 48.9 & 38.1 & 46.5 & 44.5 \\
        Matcher & 37.3 & 23.8 & 38.8 & 33.3 & 38.3 & 24.0 & 40.6 & 34.3 \\
        Ours & 53.4 & 36.8 & 51.2 & 47.1 & 59.3 & 39.9 & 58.0 & 52.4\\
        \bottomrule
    \end{tabular}
    \label{tab:disaid}
\end{table}

\subsubsection{Multiple Random Seeds Experiment}

\begin{figure}[!h]
    \centering
    \includegraphics[width=\linewidth]{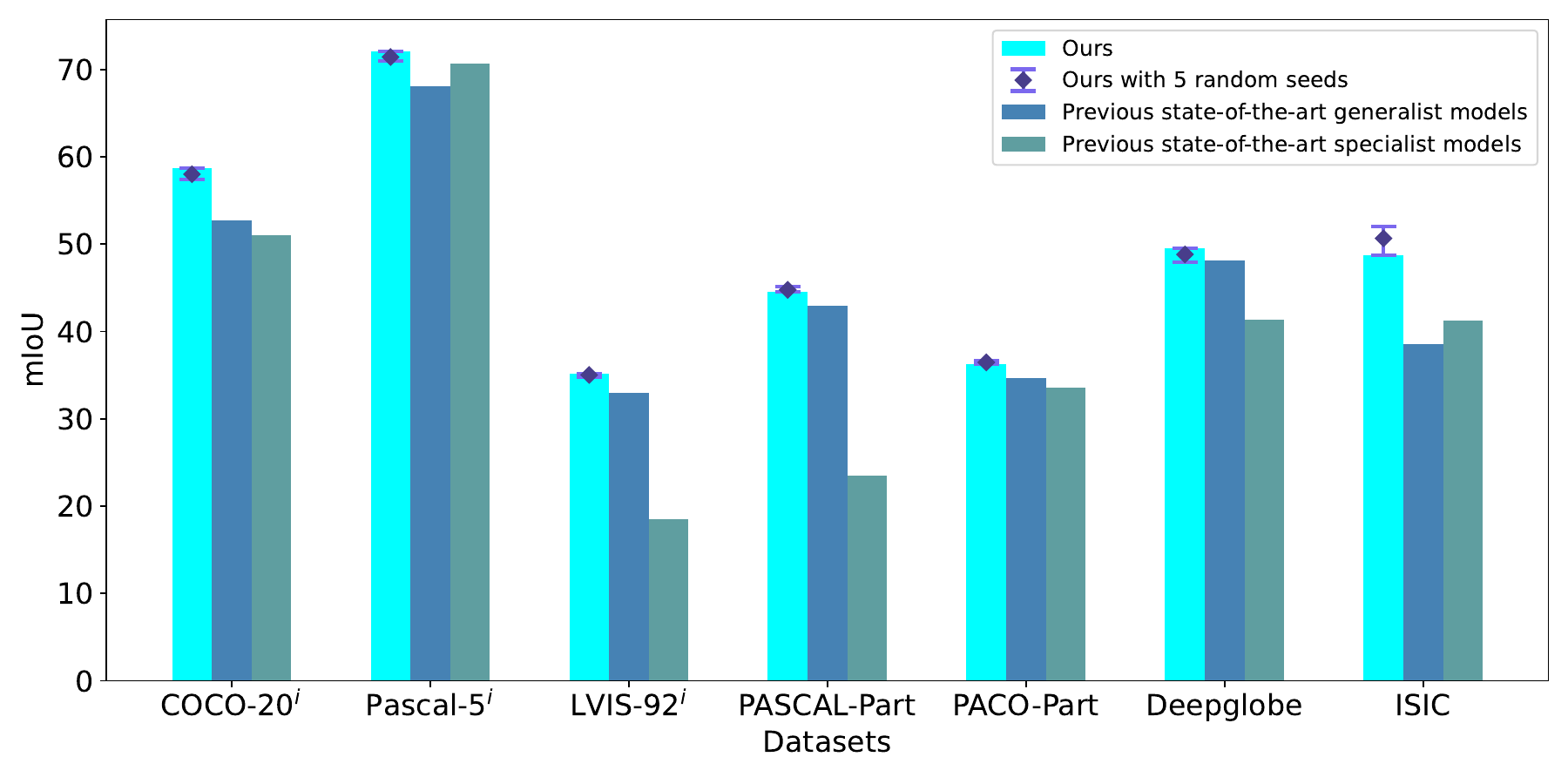}
    \caption{Results of our approach in multiple random seeds experiment. The bars in the chart represent the result under the previous standard evaluation. The error bar depicts the boundaries of our performance. }
    \label{fig:errorbar}
\end{figure}

Previous state-of-the-art methods, including both specialist methods and generalist methods, typically do not conduct multiple random seed experiments to evaluate the robustness. 
To demonstrate the robustness of our approach, we randomly set 5 different random seeds and conducted the experiments on the datasets that were not fully evaluated in the standard evaluation. 
As shown in Fig.~\ref{fig:errorbar}, despite variations in random seeds, our approach consistently exhibits better performance compared to previous methods that were not evaluated with random seeds. 

\subsection{Additional Ablation Study}

\subsubsection{Ablation Study of Pivots for Positive Gating}

We apply both $s_{mid}$ and $S^-_{mean}$ as the pivots for Positive Gating in Sec.~\ref{sec:posjudge}, aiming to leverage both the pivots from the $S^+_{mean}$ itself and the negative similarity. 
As shown in Tab.~\ref{tab:abpjp}, combining these two pivots for Positive Gating yields a significant improvement compared to using only one pivot. 
Moreover, the combination method of $\times$ shows a 0.3\% mIoU enhancement compared to $+$. 

\begin{table}[t]
    \centering
    \caption{Ablation study of pivots and combination operations in Positive Gating. }
    \begin{tabular}{cc|c|c}
    \toprule
         \multicolumn{2}{c|}{Pivots} & \multirow{2}{*}{Combination} & \multirow{2}{*}{COCO-20\textsuperscript{i}}  \\
         $s_{mid}$ & $S^-_{mean}$ & & \\
         \midrule
         \checkmark & & & 44.5 \\
          & \checkmark & & 51.0 \\
        \checkmark & \checkmark & $+$ & 56.8 \\
        \checkmark & \checkmark & $\times$ & 57.1 \\
        \bottomrule
    \end{tabular}
    \label{tab:abpjp}
\end{table}

\subsubsection{Ablation Study of Other Strategies for Positive Gating}

\begin{table}[!t]
    \centering
    \caption{Ablation study of different strategies for Positive Gating of the masks. }
    \begin{tabular}{c|cccc}
    \toprule
         Strategies & COCO-20\textsuperscript{i} & LVIS-92\textsuperscript{i} & PASCAL-Part & PACO-Part  \\
        \midrule
         Union & 59.4 & 36.1 & 40.1 & 31.6 \\
         Mask Growth & 58.7 & 35.2 & 44.5 & 36.3 \\
         \bottomrule
    \end{tabular}
    \label{tab:abspj}
\end{table}

In Sec.~\ref{sec:posjudge} and Sec.~\ref{sec:alg}, we introduce the Mask Growth algorithm as our strategy for judging whether the mask is positive. 
We compare the strategy to separately judging each mask in Tab.~\ref{tab:abpj} and judging the union mask of each cluster in Tab.~\ref{tab:abspj}. 
While simply judging the union mask shows better performance on COCO-20\textsuperscript{i} and LVIS-92\textsuperscript{i} that require complete coverage of objects, its performance on One-shot Part Segmentation has a significant decline. 
Considering the generalizability of our approach, we select the Mask Growth algorithm as our strategy. 

\subsection{Additional Qualitative Analysis}

\begin{figure}
    \centering
    \includegraphics[width=\linewidth]{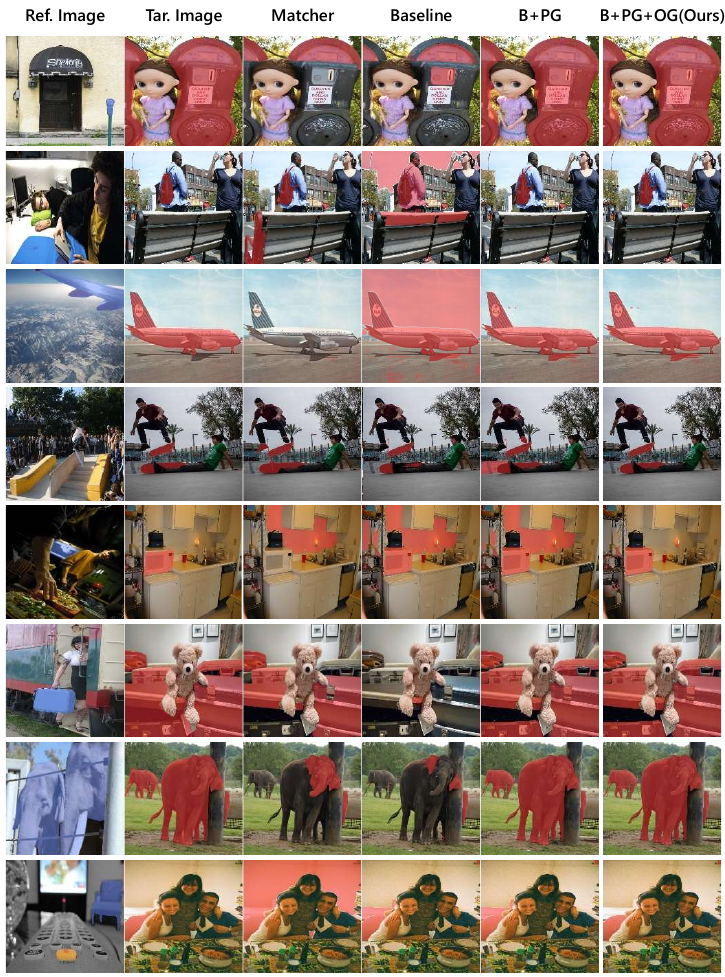}
    \caption{More Qualitative results on COCO-20\textsuperscript{i} for comparison among Matcher, Baseline, B+PG, B+PG+OG. }
    \label{fig:ap_qual}
\end{figure}

\begin{figure}
    \centering
    \includegraphics[width=\linewidth]{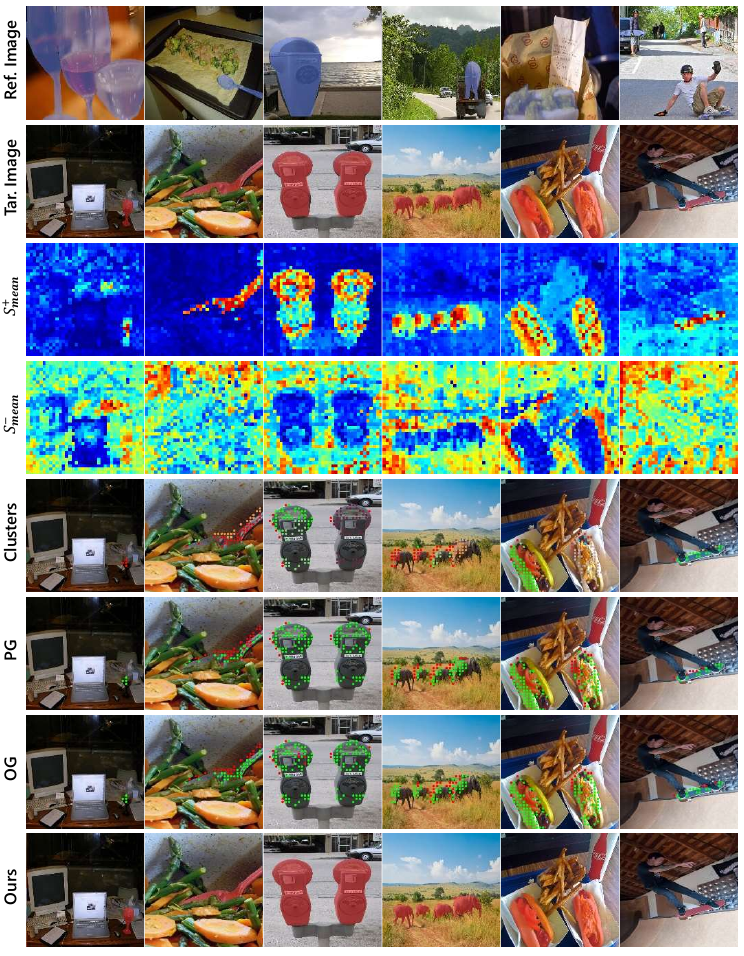}
    \caption{Qualitative analysis of the contents in gating. Different colors of points in the images in column ``Clusters" represent different clusters. The green points in images in columns ``PG" and ``OG" denote the points satisfying the gating criteria, while the red points denote those not satisfying. }
    \label{fig:ap_qualmid}
\end{figure}

\begin{figure}
    \centering
    \includegraphics[width=\linewidth]{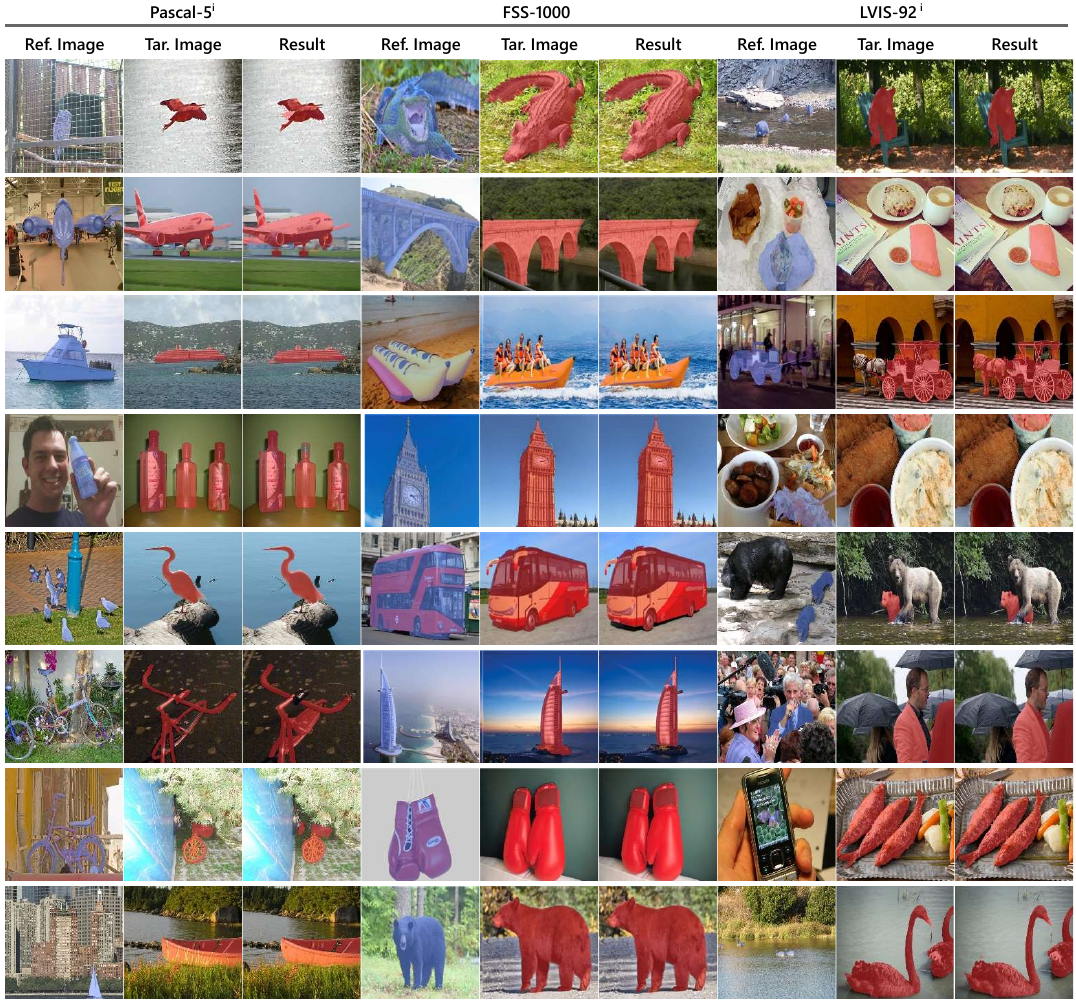}
    \caption{Qualitative analysis of the results on Pascal-5\textsuperscript{i}, FSS-1000, and LVIS-92\textsuperscript{i}. }
    \label{fig:ap_qualstd}
\end{figure}

\begin{figure}
    \centering
    \includegraphics[width=\linewidth]{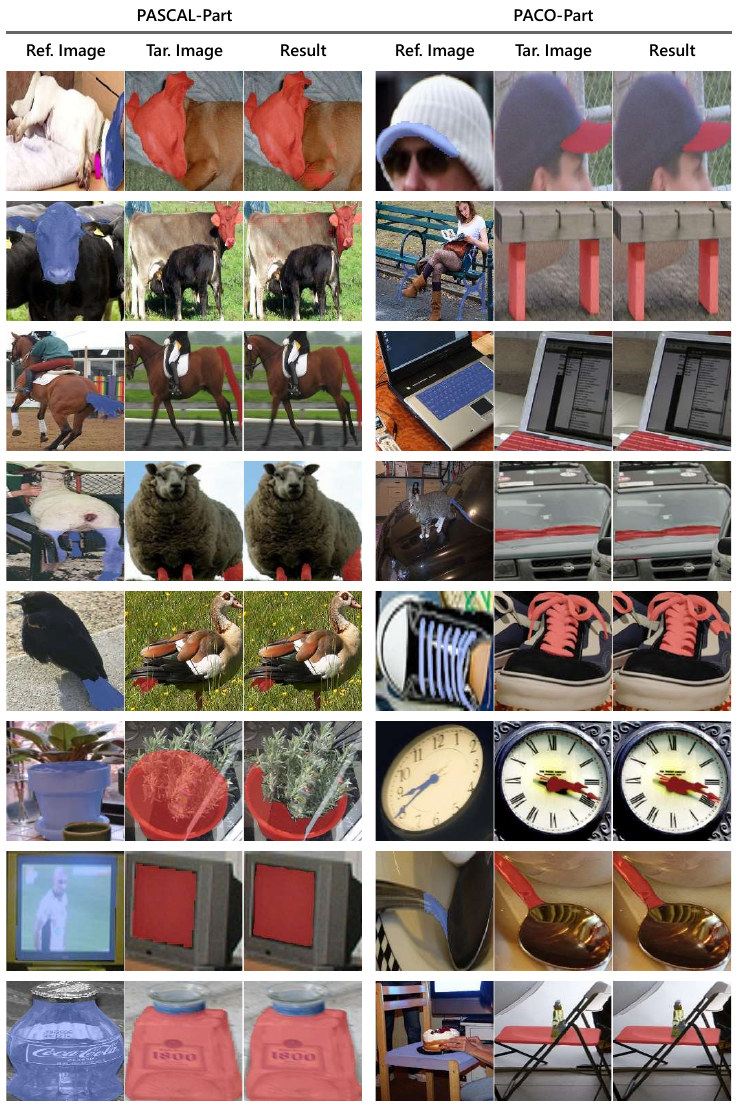}
    \caption{Qualitative analysis of the results on PASCAL-Part and PACO-Part. }
    \label{fig:ap_qualpart}
\end{figure}

\begin{figure}
    \centering
    \includegraphics[width=\linewidth]{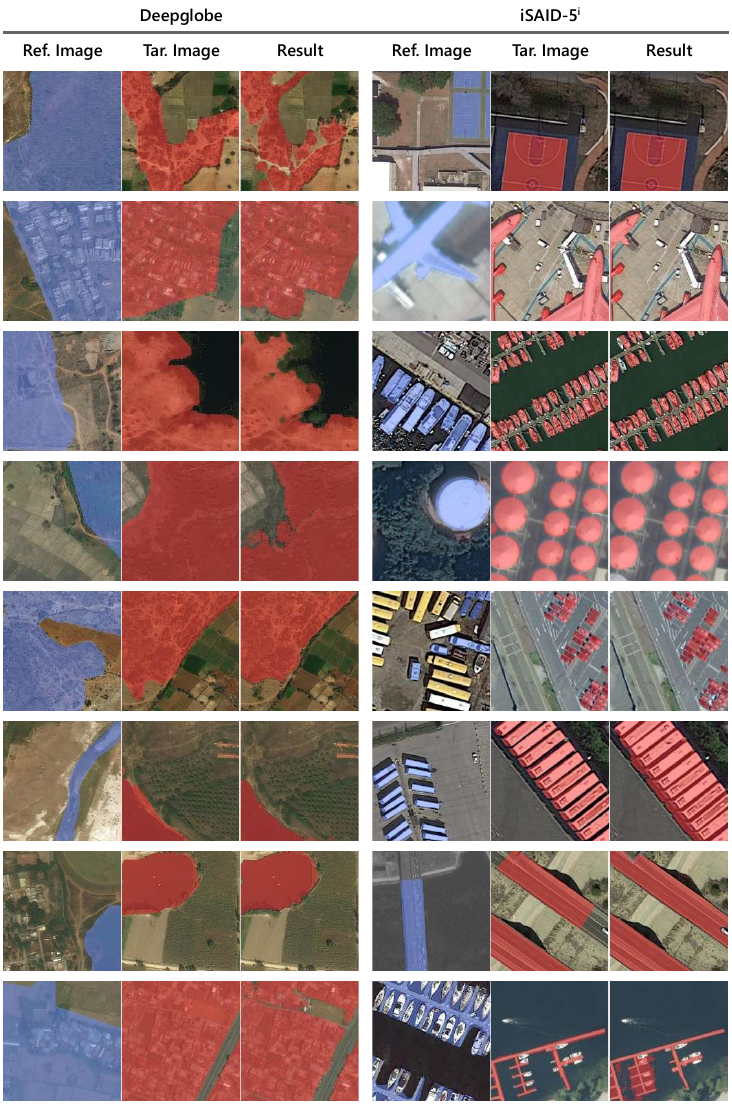}
    \caption{Qualitative analysis of the results on Deepglobe and iSAID-5\textsuperscript{i}. }
    \label{fig:ap_qualcd}
\end{figure}

We conduct additional qualitative analysis to better present the result of our approach. 
Fig.~\ref{fig:ap_qual} further compare the Matcher, Baseline, B+PJ, and B+PJ+OJ~(Ours) following Sec.~\ref{sec:qual}. 
Fig.~\ref{fig:ap_qualmid} illustrate the intermediate contents in the Post Gating. 
Moreover, we provide additional visualization results of standard FSS in Fig.~\ref{fig:ap_qualstd}, One-shot Part Segmentation in Fig.~\ref{fig:ap_qualpart}, and Cross Domain FSS in Fig.~\ref{fig:ap_qualcd}. 
These qualitative results demonstrate the effectiveness of our approach. Notably, some of the results are even better than the corresponding annotations. 

\clearpage
\section*{NeurIPS Paper Checklist}

\begin{enumerate}

\item {\bf Claims}
    \item[] Question: Do the main claims made in the abstract and introduction accurately reflect the paper's contributions and scope?
    \item[] Answer: \answerYes{} 
    \item[] Justification: Our main claims made in the abstract and the last two paragraphs of the introduction accurately reflect the paper's contributions and scope. 
    \item[] Guidelines:
    \begin{itemize}
        \item The answer NA means that the abstract and introduction do not include the claims made in the paper.
        \item The abstract and/or introduction should clearly state the claims made, including the contributions made in the paper and important assumptions and limitations. A No or NA answer to this question will not be perceived well by the reviewers. 
        \item The claims made should match theoretical and experimental results, and reflect how much the results can be expected to generalize to other settings. 
        \item It is fine to include aspirational goals as motivation as long as it is clear that these goals are not attained by the paper. 
    \end{itemize}

\item {\bf Limitations}
    \item[] Question: Does the paper discuss the limitations of the work performed by the authors?
    \item[] Answer: \answerYes{} 
    \item[] Justification: We discuss the limitations in the appendix. 
    \item[] Guidelines:
    \begin{itemize}
        \item The answer NA means that the paper has no limitation while the answer No means that the paper has limitations, but those are not discussed in the paper. 
        \item The authors are encouraged to create a separate "Limitations" section in their paper.
        \item The paper should point out any strong assumptions and how robust the results are to violations of these assumptions (e.g., independence assumptions, noiseless settings, model well-specification, asymptotic approximations only holding locally). The authors should reflect on how these assumptions might be violated in practice and what the implications would be.
        \item The authors should reflect on the scope of the claims made, e.g., if the approach was only tested on a few datasets or with a few runs. In general, empirical results often depend on implicit assumptions, which should be articulated.
        \item The authors should reflect on the factors that influence the performance of the approach. For example, a facial recognition algorithm may perform poorly when image resolution is low or images are taken in low lighting. Or a speech-to-text system might not be used reliably to provide closed captions for online lectures because it fails to handle technical jargon.
        \item The authors should discuss the computational efficiency of the proposed algorithms and how they scale with dataset size.
        \item If applicable, the authors should discuss possible limitations of their approach to address problems of privacy and fairness.
        \item While the authors might fear that complete honesty about limitations might be used by reviewers as grounds for rejection, a worse outcome might be that reviewers discover limitations that aren't acknowledged in the paper. The authors should use their best judgment and recognize that individual actions in favor of transparency play an important role in developing norms that preserve the integrity of the community. Reviewers will be specifically instructed to not penalize honesty concerning limitations.
    \end{itemize}

\item {\bf Theory Assumptions and Proofs}
    \item[] Question: For each theoretical result, does the paper provide the full set of assumptions and a complete (and correct) proof?
    \item[] Answer: \answerNA{} 
    \item[] Justification: We do not introduce theory assumptions and proofs. 
    \item[] Guidelines:
    \begin{itemize}
        \item The answer NA means that the paper does not include theoretical results. 
        \item All the theorems, formulas, and proofs in the paper should be numbered and cross-referenced.
        \item All assumptions should be clearly stated or referenced in the statement of any theorems.
        \item The proofs can either appear in the main paper or the supplemental material, but if they appear in the supplemental material, the authors are encouraged to provide a short proof sketch to provide intuition. 
        \item Inversely, any informal proof provided in the core of the paper should be complemented by formal proofs provided in appendix or supplemental material.
        \item Theorems and Lemmas that the proof relies upon should be properly referenced. 
    \end{itemize}

    \item {\bf Experimental Result Reproducibility}
    \item[] Question: Does the paper fully disclose all the information needed to reproduce the main experimental results of the paper to the extent that it affects the main claims and/or conclusions of the paper (regardless of whether the code and data are provided or not)?
    \item[] Answer: \answerYes{} 
    \item[] Justification: Our paper fully discloses all the information needed to reproduce the main experimental results of the paper. 
    \item[] Guidelines:
    \begin{itemize}
        \item The answer NA means that the paper does not include experiments.
        \item If the paper includes experiments, a No answer to this question will not be perceived well by the reviewers: Making the paper reproducible is important, regardless of whether the code and data are provided or not.
        \item If the contribution is a dataset and/or model, the authors should describe the steps taken to make their results reproducible or verifiable. 
        \item Depending on the contribution, reproducibility can be accomplished in various ways. For example, if the contribution is a novel architecture, describing the architecture fully might suffice, or if the contribution is a specific model and empirical evaluation, it may be necessary to either make it possible for others to replicate the model with the same dataset, or provide access to the model. In general. releasing code and data is often one good way to accomplish this, but reproducibility can also be provided via detailed instructions for how to replicate the results, access to a hosted model (e.g., in the case of a large language model), releasing of a model checkpoint, or other means that are appropriate to the research performed.
        \item While NeurIPS does not require releasing code, the conference does require all submissions to provide some reasonable avenue for reproducibility, which may depend on the nature of the contribution. For example
        \begin{enumerate}
            \item If the contribution is primarily a new algorithm, the paper should make it clear how to reproduce that algorithm.
            \item If the contribution is primarily a new model architecture, the paper should describe the architecture clearly and fully.
            \item If the contribution is a new model (e.g., a large language model), then there should either be a way to access this model for reproducing the results or a way to reproduce the model (e.g., with an open-source dataset or instructions for how to construct the dataset).
            \item We recognize that reproducibility may be tricky in some cases, in which case authors are welcome to describe the particular way they provide for reproducibility. In the case of closed-source models, it may be that access to the model is limited in some way (e.g., to registered users), but it should be possible for other researchers to have some path to reproducing or verifying the results.
        \end{enumerate}
    \end{itemize}

\item {\bf Open access to data and code}
    \item[] Question: Does the paper provide open access to the data and code, with sufficient instructions to faithfully reproduce the main experimental results, as described in supplemental material?
    \item[] Answer: \answerYes{} 
    \item[] Justification: Our code and instructions are included in the supplementary material. 
    The data we use for the experiments are all from open-access datasets. 
    \item[] Guidelines:
    \begin{itemize}
        \item The answer NA means that paper does not include experiments requiring code.
        \item Please see the NeurIPS code and data submission guidelines (\url{https://nips.cc/public/guides/CodeSubmissionPolicy}) for more details.
        \item While we encourage the release of code and data, we understand that this might not be possible, so “No” is an acceptable answer. Papers cannot be rejected simply for not including code, unless this is central to the contribution (e.g., for a new open-source benchmark).
        \item The instructions should contain the exact command and environment needed to run to reproduce the results. See the NeurIPS code and data submission guidelines (\url{https://nips.cc/public/guides/CodeSubmissionPolicy}) for more details.
        \item The authors should provide instructions on data access and preparation, including how to access the raw data, preprocessed data, intermediate data, and generated data, etc.
        \item The authors should provide scripts to reproduce all experimental results for the new proposed method and baselines. If only a subset of experiments are reproducible, they should state which ones are omitted from the script and why.
        \item At submission time, to preserve anonymity, the authors should release anonymized versions (if applicable).
        \item Providing as much information as possible in supplemental material (appended to the paper) is recommended, but including URLs to data and code is permitted.
    \end{itemize}

\item {\bf Experimental Setting/Details}
    \item[] Question: Does the paper specify all the training and test details (e.g., data splits, hyperparameters, how they were chosen, type of optimizer, etc.) necessary to understand the results?
    \item[] Answer: \answerYes{} 
    \item[] Justification: Our paper specifies all the test details for our training-free approach in the section of experiments. 
    \item[] Guidelines:
    \begin{itemize}
        \item The answer NA means that the paper does not include experiments.
        \item The experimental setting should be presented in the core of the paper to a level of detail that is necessary to appreciate the results and make sense of them.
        \item The full details can be provided either with the code, in appendix, or as supplemental material.
    \end{itemize}

\item {\bf Experiment Statistical Significance}
    \item[] Question: Does the paper report error bars suitably and correctly defined or other appropriate information about the statistical significance of the experiments?
    \item[] Answer: \answerYes{} 
    \item[] Justification: We set 5 random seeds to evaluate the large datasets and evaluate all samples of the small datasets in the appendix. 
    \item[] Guidelines:
    \begin{itemize}
        \item The answer NA means that the paper does not include experiments.
        \item The authors should answer "Yes" if the results are accompanied by error bars, confidence intervals, or statistical significance tests, at least for the experiments that support the main claims of the paper.
        \item The factors of variability that the error bars are capturing should be clearly stated (for example, train/test split, initialization, random drawing of some parameter, or overall run with given experimental conditions).
        \item The method for calculating the error bars should be explained (closed form formula, call to a library function, bootstrap, etc.)
        \item The assumptions made should be given (e.g., Normally distributed errors).
        \item It should be clear whether the error bar is the standard deviation or the standard error of the mean.
        \item It is OK to report 1-sigma error bars, but one should state it. The authors should preferably report a 2-sigma error bar than state that they have a 96\% CI, if the hypothesis of Normality of errors is not verified.
        \item For asymmetric distributions, the authors should be careful not to show in tables or figures symmetric error bars that would yield results that are out of range (e.g. negative error rates).
        \item If error bars are reported in tables or plots, The authors should explain in the text how they were calculated and reference the corresponding figures or tables in the text.
    \end{itemize}

\item {\bf Experiments Compute Resources}
    \item[] Question: For each experiment, does the paper provide sufficient information on the computer resources (type of compute workers, memory, time of execution) needed to reproduce the experiments?
    \item[] Answer: \answerYes{} 
    \item[] Justification: Computer resources are described in Implementation Details of the experiment section. 
    \item[] Guidelines:
    \begin{itemize}
        \item The answer NA means that the paper does not include experiments.
        \item The paper should indicate the type of compute workers CPU or GPU, internal cluster, or cloud provider, including relevant memory and storage.
        \item The paper should provide the amount of compute required for each of the individual experimental runs as well as estimate the total compute. 
        \item The paper should disclose whether the full research project required more compute than the experiments reported in the paper (e.g., preliminary or failed experiments that didn't make it into the paper). 
    \end{itemize}
    
\item {\bf Code Of Ethics}
    \item[] Question: Does the research conducted in the paper conform, in every respect, with the NeurIPS Code of Ethics \url{https://neurips.cc/public/EthicsGuidelines}?
    \item[] Answer: \answerYes{} 
    \item[] Justification: Our research conforms with the NeurIPS Code of Ethics in every respect. 
    \item[] Guidelines:
    \begin{itemize}
        \item The answer NA means that the authors have not reviewed the NeurIPS Code of Ethics.
        \item If the authors answer No, they should explain the special circumstances that require a deviation from the Code of Ethics.
        \item The authors should make sure to preserve anonymity (e.g., if there is a special consideration due to laws or regulations in their jurisdiction).
    \end{itemize}

\item {\bf Broader Impacts}
    \item[] Question: Does the paper discuss both potential positive societal impacts and negative societal impacts of the work performed?
    \item[] Answer: \answerYes{} 
    \item[] Justification: We discuss both potential positive societal impacts and negative societal impacts of our work in the appendix. 
    \item[] Guidelines:
    \begin{itemize}
        \item The answer NA means that there is no societal impact of the work performed.
        \item If the authors answer NA or No, they should explain why their work has no societal impact or why the paper does not address societal impact.
        \item Examples of negative societal impacts include potential malicious or unintended uses (e.g., disinformation, generating fake profiles, surveillance), fairness considerations (e.g., deployment of technologies that could make decisions that unfairly impact specific groups), privacy considerations, and security considerations.
        \item The conference expects that many papers will be foundational research and not tied to particular applications, let alone deployments. However, if there is a direct path to any negative applications, the authors should point it out. For example, it is legitimate to point out that an improvement in the quality of generative models could be used to generate deepfakes for disinformation. On the other hand, it is not needed to point out that a generic algorithm for optimizing neural networks could enable people to train models that generate Deepfakes faster.
        \item The authors should consider possible harms that could arise when the technology is being used as intended and functioning correctly, harms that could arise when the technology is being used as intended but gives incorrect results, and harms following from (intentional or unintentional) misuse of the technology.
        \item If there are negative societal impacts, the authors could also discuss possible mitigation strategies (e.g., gated release of models, providing defenses in addition to attacks, mechanisms for monitoring misuse, mechanisms to monitor how a system learns from feedback over time, improving the efficiency and accessibility of ML).
    \end{itemize}
    
\item {\bf Safeguards}
    \item[] Question: Does the paper describe safeguards that have been put in place for responsible release of data or models that have a high risk for misuse (e.g., pretrained language models, image generators, or scraped datasets)?
    \item[] Answer: \answerNA{} 
    \item[] Justification: Our paper has no such risks. 
    \item[] Guidelines:
    \begin{itemize}
        \item The answer NA means that the paper poses no such risks.
        \item Released models that have a high risk for misuse or dual-use should be released with necessary safeguards to allow for controlled use of the model, for example by requiring that users adhere to usage guidelines or restrictions to access the model or implementing safety filters. 
        \item Datasets that have been scraped from the Internet could pose safety risks. The authors should describe how they avoided releasing unsafe images.
        \item We recognize that providing effective safeguards is challenging, and many papers do not require this, but we encourage authors to take this into account and make a best faith effort.
    \end{itemize}

\item {\bf Licenses for existing assets}
    \item[] Question: Are the creators or original owners of assets (e.g., code, data, models), used in the paper, properly credited and are the license and terms of use explicitly mentioned and properly respected?
    \item[] Answer: \answerYes{} 
    \item[] Justification: The existing assets used in our paper, i.e., SAM and DINOv2, are released on GitHub under Apache License 2.0. 
    \item[] Guidelines:
    \begin{itemize}
        \item The answer NA means that the paper does not use existing assets.
        \item The authors should cite the original paper that produced the code package or dataset.
        \item The authors should state which version of the asset is used and, if possible, include a URL.
        \item The name of the license (e.g., CC-BY 4.0) should be included for each asset.
        \item For scraped data from a particular source (e.g., website), the copyright and terms of service of that source should be provided.
        \item If assets are released, the license, copyright information, and terms of use in the package should be provided. For popular datasets, \url{paperswithcode.com/datasets} has curated licenses for some datasets. Their licensing guide can help determine the license of a dataset.
        \item For existing datasets that are re-packaged, both the original license and the license of the derived asset (if it has changed) should be provided.
        \item If this information is not available online, the authors are encouraged to reach out to the asset's creators.
    \end{itemize}

\item {\bf New Assets}
    \item[] Question: Are new assets introduced in the paper well documented and is the documentation provided alongside the assets?
    \item[] Answer: \answerYes{} 
    \item[] Justification: Our new assets introduced in the paper are well documented in the supplementary material. 
    \item[] Guidelines:
    \begin{itemize}
        \item The answer NA means that the paper does not release new assets.
        \item Researchers should communicate the details of the dataset/code/model as part of their submissions via structured templates. This includes details about training, license, limitations, etc. 
        \item The paper should discuss whether and how consent was obtained from people whose asset is used.
        \item At submission time, remember to anonymize your assets (if applicable). You can either create an anonymized URL or include an anonymized zip file.
    \end{itemize}

\item {\bf Crowdsourcing and Research with Human Subjects}
    \item[] Question: For crowdsourcing experiments and research with human subjects, does the paper include the full text of instructions given to participants and screenshots, if applicable, as well as details about compensation (if any)? 
    \item[] Answer: \answerNA{} 
    \item[] Justification: There is no crowdsourcing and research with human subjects in our paper. 
    \item[] Guidelines:
    \begin{itemize}
        \item The answer NA means that the paper does not involve crowdsourcing nor research with human subjects.
        \item Including this information in the supplemental material is fine, but if the main contribution of the paper involves human subjects, then as much detail as possible should be included in the main paper. 
        \item According to the NeurIPS Code of Ethics, workers involved in data collection, curation, or other labor should be paid at least the minimum wage in the country of the data collector. 
    \end{itemize}

\item {\bf Institutional Review Board (IRB) Approvals or Equivalent for Research with Human Subjects}
    \item[] Question: Does the paper describe potential risks incurred by study participants, whether such risks were disclosed to the subjects, and whether Institutional Review Board (IRB) approvals (or an equivalent approval/review based on the requirements of your country or institution) were obtained?
    \item[] Answer: \answerNA{} 
    \item[] Justification: Our paper does not involve crowdsourcing or research with human subjects.
    \item[] Guidelines:
    \begin{itemize}
        \item The answer NA means that the paper does not involve crowdsourcing nor research with human subjects.
        \item Depending on the country in which research is conducted, IRB approval (or equivalent) may be required for any human subjects research. If you obtained IRB approval, you should clearly state this in the paper. 
        \item We recognize that the procedures for this may vary significantly between institutions and locations, and we expect authors to adhere to the NeurIPS Code of Ethics and the guidelines for their institution. 
        \item For initial submissions, do not include any information that would break anonymity (if applicable), such as the institution conducting the review.
    \end{itemize}

\end{enumerate}

\end{document}